\documentclass[acmtog, screen, nonacm]{acmart}
\renewcommand\footnotetextcopyrightpermission[1]{}
\authorsaddresses{}
\AtBeginDocument{%
  }

\usepackage{float}
\usepackage{wrapfig}
\usepackage{multirow}
\usepackage{colortbl}
\definecolor{bestcolor}{HTML}{F9D5E9}

\definecolor{darkyellow}{RGB}{179,134,0}

\begin{document}

\title{Fishbone: From One 3D Asset to a Million Controllable Edits}

\author{Yumeng He$^{1,2}$}
\author{Xiaoying Wang$^{1}$}
\author{Peihao Li$^{3}$}
\author{Yanjia Huang$^{1}$}
\author{Joe Masterjohn$^{4}$}
\author{Jiajun Wu$^{5}$}
\author{Leonidas Guibas$^{5}$}
\author{Yin Yang$^{6}$}
\author{Ying Jiang$^{1}$}
\author{Chenfanfu Jiang$^{1}$}

\renewcommand{\shortauthors}{Yumeng He et al.}

\begin{abstract}

Large-scale controllable 3D assets are critical for computer graphics, embodied AI, robotics, and interactive content creation, yet creating diverse 3D assets remains challenging due to the high cost of manual modeling and rigging. Shape deformation provides a natural way to generate diverse variations from existing meshes, but current data-driven deformation methods often rely on sparse user inputs, limiting accurate and controllable local edits. Although parametric editing frameworks enable more structured deformation, they typically require manually designed control structures and category-specific configurations. Inspired by natural creatures, where a central spine governs global shape and cross-sectional ribs control local variation, we introduce \textbf{Fishbone}, a unified representation for constructing rib-spine control structures over general shapes, enabling controllable parametric mesh deformation, reduced-space dynamics, and animation.  More specifically, given an input mesh, we first compute a geodesic scalar field through an adaptive heat method and extract iso-contours as cross-sectional ribs. We then introduce a geometry-aware center-finding procedure to construct a smooth spine threading through the rib centers, while Gaussian-weighted skinning associates surface vertices with nearby rib and spine structures. The resulting rib-spine representation enables real-time and predictable parametric deformation, where ribs govern local shape profiles such as thickness, orientation, and cross-sectional variation, while the spine controls global bending, twisting, and stretching. In addition, the same unified control structure naturally supports reduced-space simulation and keyframe animation. We further construct the \textbf{Fishbone-136K} dataset by augmenting the Hunyuan3D dataset with the proposed rib-spine representations, and demonstrate applications in joint generation of 3D shapes and rib-spine structures for controllable 3D generation, robot-learning generalization through deformation-based data augmentation, interactive mesh editing, and agentic generation. Extensive experiments demonstrate the effectiveness, efficiency, and versatility of the proposed framework.

\end{abstract}

\newcommand\blfootnote[1]{%
  \begingroup
  \renewcommand\thefootnote{}\footnote{#1}%
  \addtocounter{footnote}{-1}%
  \endgroup
}

\begin{teaserfigure}
  \includegraphics[width=\textwidth]{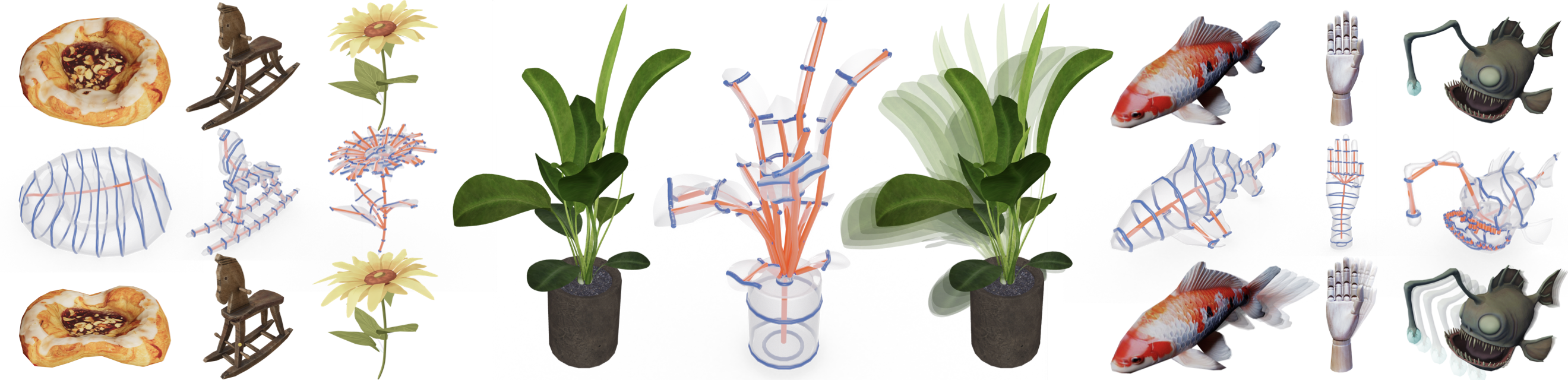}
  \caption{\textbf{Overview.} We present \textbf{Fishbone}, a novel framework that constructs a coupled rib-spine representation from an input mesh and defines rib- and spine-driven deformation primitives. The proposed representation enables diverse and controllable shape variations, spanning local cross-sectional edits as well as global bending, twisting, and stretching, all driven by intuitive low-dimensional parameters. Moreover, the rib-spine representation naturally supports real-time reduced-space simulation and keyframe animation.}
  \Description{Teaser figure showing the Fishbone control structure (rib polylines, closed loops on solid parts and open polylines on thin shells, with the spine curve threaded through their centers) overlaid on an input mesh, alongside a gallery of diverse shape variations generated by composing rib- and spine-driven deformation primitives. }
  \label{fig:teaser}
\end{teaserfigure}

\maketitle
\blfootnote{
    $^1$UCLA, 
    $^2$USC, 
    $^3$UC Berkeley, 
    $^4$TRI, 
    $^5$Stanford, 
    $^6$Utah \\
    Core contributors: Yumeng He, Xiaoying Wang, Ying Jiang
}

\section{Introduction}

The availability of large, diverse 3D shape datasets is a critical bottleneck across computer graphics, vision, and robotics.
Training robust 3D generative models~\cite{poole2022dreamfusion, nichol2022point, jun2023shap, xiang2024structured, liu2024part123}, learning dexterous manipulation policies~\cite{peng2018sim, levine2016end, akkaya2019solving}, and building sim-to-real pipelines~\cite{tobin2017domain, sadeghi2016cad2rl} all demand vast collections of geometrically varied yet plausible meshes. While 3D scanning and artist-authored assets enable the reconstruction of real-world 3D geometry and the creation of virtual content, modeling unseen occluded regions in real-world scenarios is fundamentally limited by missing image or scan observations. Generating free-form objects beyond man-made shapes in virtual environments typically requires expert-level modeling and sculpting. A natural solution is to leverage shape deformation to generate controllable variations of existing shapes, thereby alleviating the need for manual modeling \cite{tan2018mesh}. 

To achieve controllable mesh deformation, prior work explores cage-based coordinates~\cite{ju2023mean}, skeleton-driven animation~\cite{baran2007automatic}, and free-form deformation lattices~\cite{sederberg1986free, coquillart1990extended}. While these methods offer precise, repeatable control, they require substantial manual setup, such as rigging skeletons, designing cages, or placing lattices that do not scale to large collections of heterogeneous assets. Learning-based methods, in contrast, predict deformation fields from general sparse user inputs, enabling deformation across a broader range of shapes without additional manual setup. Previous approaches use sketch strokes~\cite{igarashi2006teddy} or projected sparse points~\cite{han2017deepsketch2face} as conditioning to provide an intuitive interface, but these 2D control signals, due to depth uncertainty and view-dependent artifacts, limit precision on fine-grained local geometry. Recent methods that lift control into 3D, through handle points~\cite{yumer2015semantic}, neural cages~\cite{yifan2020neural}, or learned flow fields~\cite{zhu2024controllable, hanocka2018alignet, yumer2016learning}, recover spatial accuracy but they are often category-specific, require dedicated training data, and degrade on out-of-distribution shapes. In addition, sparse user-defined control signals fail to establish dense, one-to-one correspondence with the geometry, limiting precise and predictable local edits. In contrast, parametric modeling encodes shape variation through explicit, low-dimensional parameters tied to geometric structure~\cite{loper2023smpl, blanz2023morphable, wu2021deepcad}, enabling systematic, fine-grained control over shape. However, these approaches rely on well-defined parametric structures and clean topology that work for man-made shapes but break down on free-form geometry. This raises the following question: can we construct a geometry-derived, low-dimensional control representation for general shapes that enables predictable parametric deformation without manual rigging, category-specific training, or predefined shape structure?

Inspired by natural creatures such as plants, animals, and organic tissues, where a central spine governs global extent and orientation while cross-sectional ribs control local width and surface variation, we extract an analogous control structure directly from the mesh. Specifically, we design a spine curve threading through the volumetric center together with surface-aligned cross-sectional ribs sampled at successive geodesic levels, forming a compact and interpretable handle representation that naturally adapts to diverse triangle-mesh geometries without category-specific supervision. In this paper, we present \textbf{Fishbone}, a novel framework for constructing rib-spine control structures for diverse and controllable mesh deformation. Specifically, we introduce a compact 3D rib-spine representation, where ribs encode local shape profiles such as thickness, orientation, and cross-sectional variation, while the spine governs the global shape trajectory and deformation. Given an input mesh with one or multiple parts, we first compute a geodesic scalar field using an adaptive heat method. Iso-contours of this field are extracted as cross-sectional ribs, represented as closed loops on solid meshes and open polylines on thin shells or single-sided sheets. A geometry-aware center-finding algorithm then constructs a smooth \emph{spine} curve through the rib centers. We further compute per-vertex skinning weights that associate each surface point with nearby rib and spine segments through Gaussian radial basis functions. The resulting explicit rib-spine representation establishes a structured local-global control representation, where local rib-driven deformations include uniform or anisotropic scaling, translation, rotation, local dragging, and cross-sectional reshaping, while global spine-driven deformations support intuitive twisting, bending, and stretching. This parameterization enables intuitive and accurate edits across diverse geometries. In addition, the generated rib-spine structure also supports reduced-space simulation and keyframe animation, enabling efficient and visually plausible motions through the same unified control representation. We demonstrate that the proposed rib-spine representation enables more accurate and controllable 3D editing and generation than the baselines. Beyond interactive editing, the explicit parametric control structure also supports scalable 3D asset augmentation for downstream applications such as robot learning and agentic mesh generation. To summarize, our contributions are:

\vspace{-10pt}

\begin{itemize}
    \item We introduce a unified \textbf{Fishbone} representation that couples geometry-derived ribs, a smooth spine, and Gaussian-weighted skinning, enabling intuitive and efficient local rib-driven and global spine-driven parametric deformations. We further explore the proposed representation for reduced-space dynamics, including physics-aware simulation and keyframe animation through the same unified rib-spine structure.
    
    \item We propose a novel mesh-to-\textbf{Fishbone} framework that automatically extracts rib-spine structures from general meshes without manual rigging or category-specific training. Building on this pipeline, we construct the \textbf{Fishbone-136K} dataset containing part-level meshes together with corresponding ribs, spines, and skinning structures. We further explore an autoregressive model for feed-forward joint generation of 3D shapes and rib-spine control structures.

    \item Versatile applications and comprehensive evaluations, including 3D generation, interactive editing, robot learning, and agentic generation, demonstrate the effectiveness and efficiency of the proposed method.
\end{itemize}

\section{Related Work}

\subsection{3D Generation and Editing}
Most 3D shapes in product design are created through parametric and procedural modeling, including CAD systems and constructive solid geometry, which encode geometry as operations over low-dimensional parameters~\cite{wu2021deepcad}. While precise and deterministic, these methods rely on clean control structures suited to engineered objects but struggle with free-form shapes, often requiring expert sculpting. Learning-based generative models allow users to create 3D shapes intuitively and efficiently, without additional sculpting or manual parameter tuning by exploring from autoencoders~\cite{wu2016learning} and flow-based models~\cite{yang2019pointflow} to diffusion and transformer architectures that generate assets from text or images across point clouds~\cite{luo2021diffusion}, implicit fields~\cite{jun2023shap}, meshes~\cite{liu2023meshdiffusion}, and part-aware representations~\cite{lin2025partcrafter, liu2024part123}. Nevertheless, these models remain limited by their training distributions, with brittle out-of-distribution performance and limited control over fine-grained variation. To further improve shape diversity and controllability, prior work explores explicit conditioning signals for editing, including keypoints, trajectories, primitives~\cite{fedele2025spacecontrol}, and image or text guidance. While these approaches enable shape editing from sparse inputs, they offer limited fine-grained control. In contrast, template-based methods~\cite{tan2018mesh, loper2023smpl} rely on fixed parametric templates for deformation rather than sparse intrinsic, geometry-derived control signals. Although they enable detailed and accurate shape deformation, they typically require complex parameter tuning. Our goal is to enable accurate, continuous, and controllable shape editing for diverse 3D shapes using an automatically extracted low-dimensional control structure.

\subsection{Mesh Deformation}
Shape deformation has been formulated by equipping users with a set of handles that define a deformation space \cite{gao2023textdeformer}. Data-driven methods predict deformation fields of free-form shapes directly via neural cages~\cite{yifan2020neural}, semantic deformation flows~\cite{yumer2016learning}, part-shape alignment networks~\cite{hanocka2018alignet}, and neural generalized cylinders~\cite{zhu2024controllable}, which generalize across categories. These non-parametric approaches support arbitrary mesh deformation via latent codes or neural fields. However, neural edits are governed by learned representations whose effects on local geometry are often non-deterministic and hard to interpret, typically requiring category-specific training~\cite{mitra2014structure}. To enable accurate editing, prior work has developed parametric deformation frameworks based on energy minimization and skinning, where vertex positions are deterministically interpolated from sparse handle controls. Representative approaches include cage-based barycentric coordinates \cite{ju2023mean, joshi2007harmonic, lipman2008green}, free-form deformation lattices \cite{sederberg1986free, coquillart1990extended}, skeleton-driven linear and dual quaternion skinning \cite{magnenat1989joint, lewis2023pose, kavan2007skinning}, and as-rigid-as-possible (ARAP) deformation~\cite{sorkine2007rigid}. These methods provide precise control, but typically require user-specified handles, cages, skeletons, or other control structures, making them costly to apply at scale to heterogeneous asset collections. In contrast, \textbf{Fishbone} derives an explicit handle structure directly from each input mesh, exposing interpretable rib- and spine-driven parameters while preserving mesh connectivity and requiring no per-category training.

\subsection{Skeletal and Structural Shape Representations}
Skeletal and structural representations of 3D shapes have been studied for decades~\cite{tagliasacchi20163d}.
Curve skeletons~\cite{tagliasacchi2009curve} capture topology and coarse geometry as 1D abstractions, and can be computed via Voronoi-based thinning~\cite{foskey2001voronoi, amenta2001power}, Laplacian contraction~\cite{au2008skeleton}, and mean curvature flow~\cite{tagliasacchi2012mean}. Boundary representations (B-rep)~\cite{lee2025brepdiff} model solids through their bounding faces, edges, and vertices. Nevertheless, these representations either focus on structural abstraction or boundary geometry, and often lack a tight coupling between intrinsic structure and surface detail. Moreover, B-reps are primarily designed for man-made shapes and do not generalize well to free-form geometry. Geodesic distance computation on meshes provides a complementary surface-aware tool for geometric processing, with well-established methods such as the heat method~\cite{crane2013geodesics}, fast marching~\cite{sethian1999fast}, and exact polyhedral algorithms~\cite{surazhsky2005fast}. To better capture interior structure, meso-skeletons~\cite{huang2013l1} and medial axis transforms~\cite{choi1997mathematical} offer richer intrinsic representations, but are sensitive to surface noise and topology changes, and may become unstable or drift outside the shape boundary \cite{guo2024medial}. Inspired by the skeletal structure of creatures, where a central spine governs global structure and ribs encode local shape variations, we propose a \textbf{Fishbone} representation consisting of internal spines with surface-aligned ribs, enabling unified internal and surface-aware modeling, and further supports mesh deformation, simulation, and animation.

\subsection{Reduced Simulation and Animation}
\label{sec:related_simulation}
Reduced-order modeling enables efficient, even real-time simulation in interactive settings by projecting the full vertex state onto a low-dimensional subspace, either linear~\cite{barbivc2005real} or nonlinear~\cite{fulton2019latent}, and reconstructing motion on the surface~\cite{zong2023neural}. To construct reduced subspaces, classical modal analysis~\cite{pentland1989good} extracts dominant vibration modes of discretized elastic objects, and subsequent work extends this framework to POD-based subspace integration~\cite{barbivc2005real}, modal warping for moderate rotations~\cite{choi2005modal}, and reduced finite-element or projective integrators that retain plausible deformation at orders-of-magnitude lower cost~\cite{bouaziz2023projective}, often built on top of as-rigid-as-possible~\cite{sorkine2007rigid} or position-based full models. After obtaining subspace dynamics, they are lifted back to the full space for surface deformation~\cite{kim2011physics}. During this lift process, previous work on linear blend skinning~\cite{magnenat1989joint} computes vertex positions as weighted combinations of nearby joint transformations, dual-quaternion skinning~\cite{kavan2007skinning} alleviates volume-loss artifacts, and pose-space deformation~\cite{lewis2023pose} introduces corrective shapes to handle extreme poses. This subspace-to-surface deformation is also used in rig-based animation to offer intuitive and efficient control over surface deformation, where an authored or automatically generated~\cite{baran2007automatic} skeletal armature, together with skinning weights, drives the surface. In addition, keyframe animation uses a low-dimensional temporal control space, where sparse poses are specified, and intermediate frames are obtained via spline interpolation~\cite{kochanek1984interpolating}. However, these animation approaches typically require artists to manually define control structures, such as skeletons or keyframes~\cite{lasseter1998principles}. In contrast, our method supports both editing and simulation within a unified pipeline without requiring manual specification of the control subspace: the spine points serve as the subspace coordinates, and a precomputed skinning matrix maps the result back to the surface, with physics-based energy driving simulation and interpolation in parameter space enabling animation.

\section{Fishbone}
\label{sec:rib}

We propose \textbf{Fishbone}, a novel framework that decouples local from global shape variation through three coupled components: cross-sectional iso-geodesic contours (ribs) $\mathcal{R}$, a smooth centerline (spine) $\mathcal{S}$, and skinning weights $\mathbf{W}$, forming the representation $\mathcal{F}=(\mathcal{R},\mathcal{S},\mathbf{W})$. Ribs encode local shape profiles such as thickness, orientation, and cross-sectional variation, while the spine governs the global trajectory of the shape. This design establishes a structured local-global control representation in which local edits are applied through ribs and global deformations are driven by the spine, enabling accurate and intuitive control over diverse geometries, including organic forms, man-made objects, and free-form surfaces. Given a mesh $\mathcal{M}$ with one or multiple parts, we further introduce a mesh-to-\textbf{Fishbone} pipeline that extracts coupled rib-spine structures and corresponding skinning weights without requiring category-specific templates or manual rigging. The proposed rib-spine structure can also support shape editing, reduced-space dynamics, and keyframe animation.

\begin{figure}[h]
    \centering
    \includegraphics[width=\linewidth]{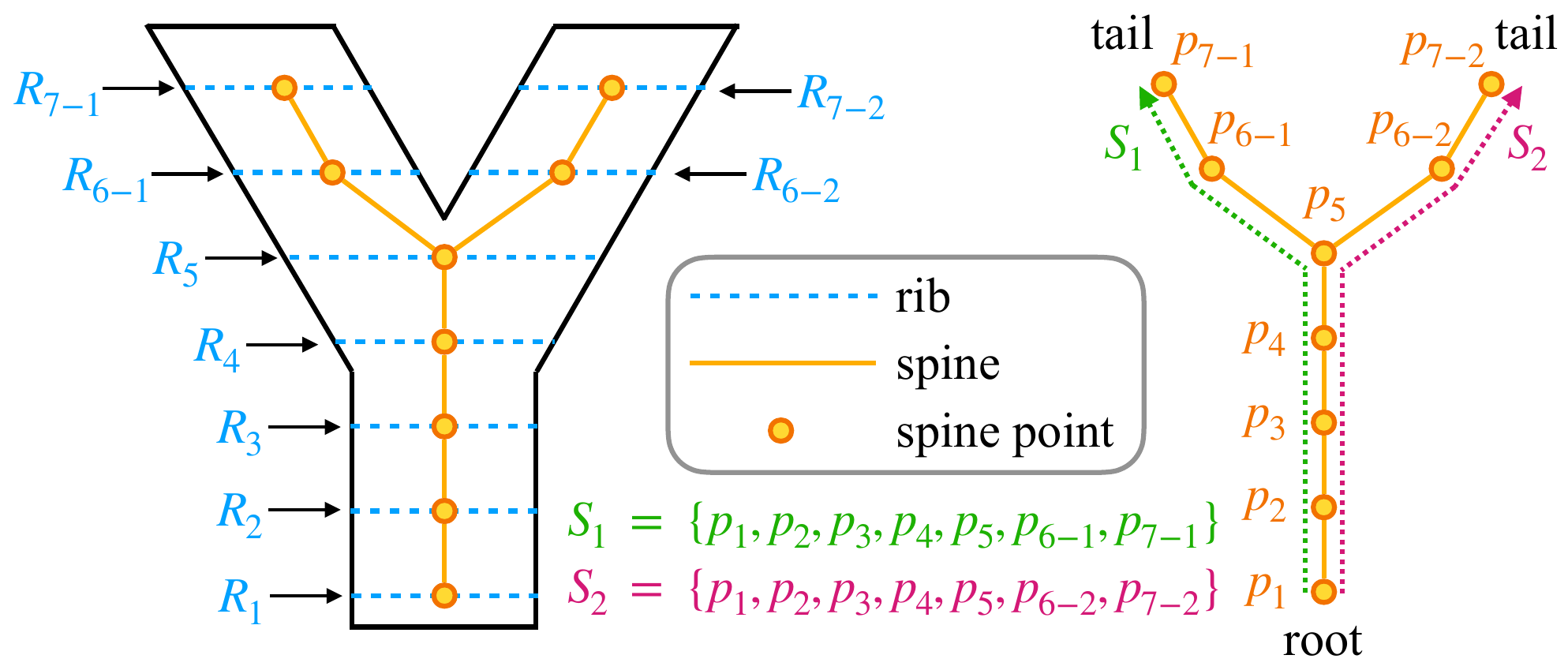}
    \caption{\textbf{Rib-spine representation and shared-node spine tree.}
    \emph{Left:} \textbf{Fishbone} represents a mesh by surface ribs $R_k$, shown as ordered iso-contour polylines, and spine key points $\mathbf{p}_k$, shown as rib centers connected into an orange spine. At branching levels, a single geodesic level may produce multiple disconnected sub-ribs, such as $R_{k,1}$ and $R_{k,2}$.
    \emph{Right:} the spine is stored as a shared-node tree: each spine $\mathbf{S}_b$ visits a subset of the same global key-point set $\{\mathbf{p}_1, \ldots, \mathbf{p}_K\}$, with junction points shared by multiple spines rather than duplicated, ensuring consistent deformation and dynamics across branches.}
    \label{fig:spine_label}
\end{figure}

\subsection{Fishbone Representation}
\label{sec:rib_intro}

Given a triangle mesh $\mathcal{M} = (\mathbf{V}, \mathbf{T})$ with vertices $\mathbf{V} \in \mathbb{R}^{N \times 3}$ and faces $\mathbf{T}$, our control structure is a tuple $\mathcal{F} = (\mathcal{R}, \mathcal{S}, \mathbf{W})$, where $\mathcal{R} = \{R_1, R_2, \ldots, R_K\}$ is a set of \emph{ribs}, $\mathcal{S}$ is the \emph{spine}, and $\mathbf{W} = (\mathbf{W}^r, \mathbf{W}^s)$ is a pair of sparse \emph{skinning weight} matrices binding mesh vertices to ribs and spine points, respectively.
\paragraph{Ribs.} Each rib $R_k$ is an ordered polyline lying on the mesh surface $R_k = (\mathbf{r}_k^1, \mathbf{r}_k^2, \ldots, \mathbf{r}_k^{M_k})$, representing the cross-section at the $k$-th geodesic level from the root. If the last point $\mathbf{r}_k^{M_k}$ is adjacent to the first point $\mathbf{r}_k^1$, then $R_k$ is treated as a closed loop. Otherwise, $R_k$ is an open polyline whose endpoints lie on the surface boundary. Ribs encode local volumetric properties of the shape: their length, orientation, and non-planarity jointly characterize how the surface expands, contracts, or twists along the structure. When an iso-geodesic level yields multiple disconnected polylines (e.g., when an arm separates from the torso or when a surface has multiple boundary components), we represent each polyline as an independent sub-rib $R_{k,l}$. To capture topology changes across levels, we maintain a parent–child tree that records how sub-ribs at level $k+1$ originate from (i.e., split from) those at level $k$.

\paragraph{Spine.}
The spine $\mathcal{S}$ consists of $K$ unique 3D spine points and a set of spines,
$\mathcal{S} = \bigl(\{\mathbf{p}_1, \ldots, \mathbf{p}_K\},\; \{\mathbf{S}_1, \mathbf{S}_2, \ldots\}\bigr)$,
where each $\mathbf{p}_k \in \mathbb{R}^3$ is the representative interior point of a rib, ordered from root to tail. Each spine $\mathbf{S}_b = (\mathbf{p}_{b,1}, \mathbf{p}_{b,2}, \ldots)$ is an ordered sequence of spine points describing a root-to-tail path through consecutive ribs. Spines share common prefixes along the trunk (e.g., $\mathbf{p}_1,\ldots,\mathbf{p}_5$) and diverge at branching points, where sub-branches follow distinct sequences (e.g., $\mathbf{p}_{6\text{-}1}, \mathbf{p}_{7\text{-}1}$ vs.\ $\mathbf{p}_{6\text{-}2}, \mathbf{p}_{7\text{-}2}$). At such Y-junctions, the branching spines meet at the same junction point and then evolve independently, forming a tree structure over a shared node set rather than duplicated branches.

\paragraph{Skinning.}
To enable smooth and consistent propagation of rib- and spine-driven edits to the mesh, we define a pair of sparse skinning matrices $\mathbf{W} = (\mathbf{W}^r, \mathbf{W}^s)$, where $\mathbf{W}^r$ binds mesh vertices to ribs and $\mathbf{W}^s$ binds them to spine points. Entries are non-zero only within each handle's local support, and each row is normalized to sum to one. Together with the per-vertex projection information (the nearest rib edge index and parametric position), $\mathbf{W}$ provides everything needed to propagate edits on the ribs or spine back to the mesh with smooth, detail-preserving falloff.

\begin{figure*}[t]
    \centering
    \includegraphics[width=\linewidth]{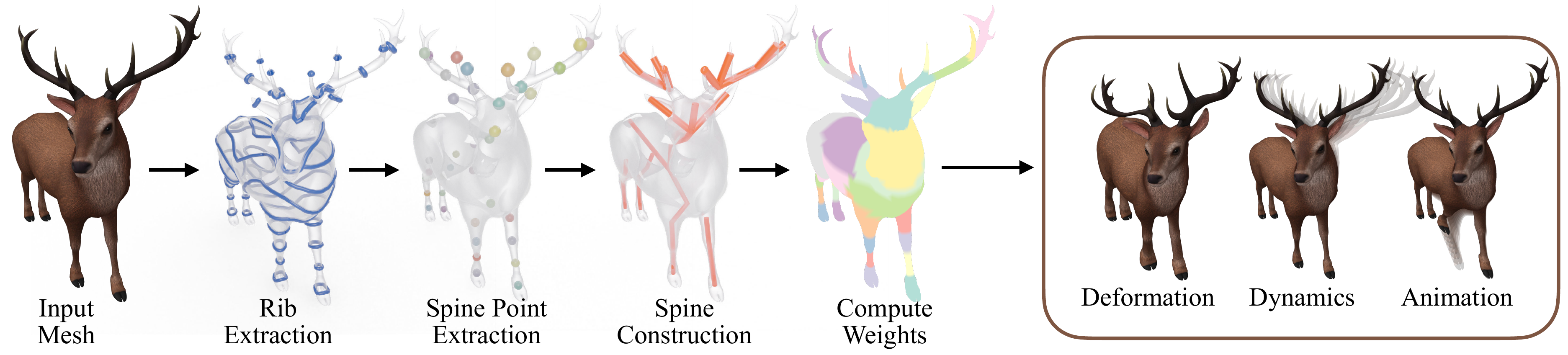}
    \caption{%
        \textbf{Pipeline Overview.} Starting from an input mesh $\mathcal{M}$ with one or multiple parts, we compute a geodesic field $\phi$ for each part using the adaptive heat method and extract ribs $\mathcal{R}$ as iso-contours of $\phi$. For each rib $R_k \in \mathcal{R}$, we select the locally flattest interior location as the representative spine point and connect them to form a branching spine $\mathcal{S}$. We then compute per-vertex skinning weights $\mathbf{W}$ that bind the mesh to the proposed rib-spine control structure. The proposed rib-spine structure further supports deformation, reduced-space dynamics, and keyframe animation within a unified representation.
    }
    \label{fig:pipeline}
\end{figure*}

\subsection{Rib Extraction} %
\label{sec:rib_gen}
To construct the rib set $\mathcal{R}=\{R_k\}$, ribs are generated independently for each segmented part of the mesh $\mathcal{M}$. For each part, we first auto-select a root region from its axis-aligned bounding box and compute a geodesic field $\phi$ using the proposed adaptive heat method. We then extract ribs as iso-contours of $\phi$ at multiple geodesic levels, where each rib is represented as an ordered polyline obtained through marching triangles and contour stitching. Aggregating ribs across all parts and levels yields the global rib set $\mathcal{R}$.

\paragraph{Automatic root-axis selection.}
To automatically initialize the root for geodesic distance computation on each mesh's segmented part, we estimate a dominant geometric direction from the part's axis-aligned bounding box. Let $\mathbf{v}_{\min}, \mathbf{v}_{\max}$ denote the bounding box corner vertices, and let $\Delta_a = (\mathbf{v}_{\max} - \mathbf{v}_{\min})_a$, $a \in \{x,y,z\}$, be the extent along each axis. We select the dominant axis $a^\star = \arg\max_a \Delta_a$. The source region is initialized from one extreme side of this axis. For upright objects, we use the minimum-$y$ face when $a^\star = y$. Otherwise, we choose the extreme face whose coordinate is closer to the world origin. The root is then defined as the $1\%$ of vertices closest to the selected face along $a^\star$. This initialization induces a consistent root-to-tail geodesic organization of ribs and can be overridden by user-specified source vertices or axis directions.

\paragraph{Geodesic field.}
Inspired by~\cite{crane2013geodesics}, we compute the geodesic field $\phi$ by solving a short-time heat diffusion process initialized at the root, and then solving a Poisson equation that reconstructs distances from the normalized heat gradient:
\begin{subequations}
\label{eq:heat}
\begin{align}
    (I - t \Delta) \mathbf{u} &= \delta_{\text{root}}, \\
    \Delta \phi &= -\nabla \cdot \bigl(\nabla \mathbf{u} / \|\nabla \mathbf{u}\|\bigr),
\end{align}
\end{subequations}
where $I$ is the identity matrix, $\Delta$ is the discrete Laplace-Beltrami operator, $t$ is the diffusion time step, and $\mathbf{u}$ denotes the diffused heat field. This yields a smooth, monotonically increasing field that is robust to local mesh noise and near-linear in the number of vertices. For parts with multiple disconnected components, we run the solver per connected component, selecting a fresh root in each component while keeping the resulting $\phi$ values on a shared scale. Small noise components therefore have tiny $\phi_{\max}$ and produce no iso-levels, whereas valid disjoint branches contribute ribs at appropriate levels.

\paragraph{Adaptive level count.}
To determine an appropriate number of ribs $K$ per part, rather than using a fixed value, we adapt $K$ according to the part’s relative extent, setting $K \propto L_p/L_o$ and restricting it to the range $[K_{\min}, K_{\max}]$, where $L_p$ and $L_o$ denote the largest bounding-box extents of the part and the full object, respectively. This yields approximately uniform inter-rib spacing across the asset while avoiding over-sampling of small parts.

\paragraph{Rib Tracing and Organization.}
After obtaining the geodesic field $\phi$, we sample $K$ geodesic levels $\{\phi_k\}_{k=1}^K$ and extract ribs $\mathcal{R}$ as iso-contours of $\phi$ at these levels. For each level $\phi_k$, contour segments are generated by intersecting the iso-level set $\phi=\phi_k$ with mesh edges whose endpoint geodesic values satisfy $\phi(\mathbf{v}_i)\le\phi_k<\phi(\mathbf{v}_j)$. The corresponding contour intersection point is computed by linear interpolation:
\begin{equation}
    \mathbf{p}_{ij}
    =
    \mathbf{v}_i
    +
    \frac{\phi_k-\phi(\mathbf{v}_i)}
    {\phi(\mathbf{v}_j)-\phi(\mathbf{v}_i)}
    (\mathbf{v}_j-\mathbf{v}_i).
\end{equation}
After extracting contour intersection points on mesh edges, we connect the resulting contour segments into ordered rib polylines by traversing neighboring triangles along each iso-level $\phi_k$. On 2-manifold regions, the resulting ribs form closed loops, while on thin-shell or single-sided regions, the contours terminate at surface boundaries and produce open polylines. However, since open polylines may arise either from genuine thin-shell boundaries or from small cracks on solid meshes, we distinguish these two cases before filtering the extracted ribs. Specifically, we classify a part as thin-shell when the fraction of edges belonging to only one triangle exceeds an empirical threshold ratio $\tau_s$. Open polylines are preserved as valid ribs only for thin-shell parts, while only closed loops are retained for solid parts.

When a single iso-level yields multiple disconnected polylines, such as at topological branch points where limbs separate from a torso or where thin-shell parts contain multiple boundary segments, we represent each polyline as an individual sub-rib $R_{k,l}$. To establish branching relationships across neighboring iso-levels, we construct a parent-child tree based on mesh topology rather than 3D proximity. Specifically, between consecutive iso-levels $\phi_{k-1}$ and $\phi_k$, we extract the corresponding face strip and perform a multi-source breadth-first search on the restricted face-adjacency graph, using sub-ribs from level $\phi_{k-1}$ as propagation sources. Each current sub-rib inherits the parent label that reaches the majority of its associated faces with the smallest hop count, with ties resolved by centroid proximity. Compared with direct connected-component matching or 3D nearest-neighbor association, this topology-aware propagation more robustly preserves branch consistency under complex geometric configurations and naturally supports downstream spine-tree construction. Fig.~\ref{fig:spine_label} illustrates the resulting branching topology. In the left panel, a single iso-contour level produces two disconnected sub-ribs ($R_{k,1}$ and $R_{k,2}$) sharing the same parent at level $k-1$, while the right panel shows the corresponding shared-node spine tree induced by this parent-child structure.

\subsection{Spine Construction}
\label{sec:spine_gen}
To construct the spine $\mathcal{S}$ as a global control structure that captures both spine positions $\mathbf{p}$ and branching paths $\mathbf{S}$ across ribs from root to tail, we first fit a local surface for each rib $R_k\in\mathcal{R}$ and then extract a representative spine point $\mathbf{p}_k$ by maximizing a geometry-aware score defined over the rib geometry. The extracted spine points are then assembled into a branching spine structure according to the parent-child relationships between neighboring ribs.

\paragraph{Geometry-aware Spine Point Extraction.}
The spine point should lie near the geometric center of the rib while remaining stable under local surface variation and branch transitions. To estimate local rib height variation and identify stable spine locations, we approximate each rib $R_k$ by a best-fit plane via principal component analysis (PCA)~\cite{abdi2010principal}, equipped with an orthonormal in-plane basis $(\mathbf e_u,\mathbf e_v)$ where $\mathbf e_v$ aligns with the global height direction. Rib vertices are then projected onto the plane to obtain 2-D coordinates $(u,v)$. Unlike a local surface tangent plane, this rib-level reference plane provides a consistent geometric ground for measuring cross-sectional flatness and height variation. We then choose the spine point $\mathbf p_k$ as the location maximizing a geometry-aware score combining local flatness, geometric centering, and parent-child continuity:
\begin{equation}
(u^\star, v^\star)
=
\arg\max_{\mathbf q\in\Omega_k}
\alpha F(\mathbf q)
+
\beta C(\mathbf q)
+
\gamma P(\mathbf q),
\end{equation}
\begin{equation}
\mathbf p_k
=
\bar{\mathbf r}_k
+
u^\star\mathbf e_u
+
v^\star\mathbf e_v,
\end{equation}
where $\bar{\mathbf r}_k$ is the centroid of rib $R_k$, $\Omega_k$ denotes the interior in-plane search region enclosed by the rib polygon, and $\alpha,\beta,\gamma \ge 0$ are weighting coefficients. $F(\mathbf q)=1-\mathrm{norm}(\tilde g(v_{\mathbf q}))$ favors locally flat regions with small height variation, $C(\mathbf q)=1-\mathrm{norm}(\|\mathbf q-\bar{\mathbf r}_k\|_2)$ encourages proximity to the rib centroid, and $P(\mathbf q)=1-\mathrm{norm}(\|\mathbf q-\mathbf p_{\mathrm{parent}}\|_2)$ promotes parent-child continuity with the parent spine point $\mathbf p_{\mathrm{parent}}$ from the BFS-based rib hierarchy of \S\ref{sec:rib_gen}, projected onto the current rib plane. $\tilde g(v)$ is a Gaussian-smoothed aggregation of the rib height derivative. For closed ribs, we uniformly sample candidate locations in the enclosed $(u,v)$ parametric domain using a regular $G\times G$ grid, while for open ribs, candidate points are directly sampled along the polyline.

\paragraph{Branching Spine Construction.}
After extracting representative spine points $\mathbf{p}_k$ from neighboring ribs, we assemble them into branch spines $\mathbf{S}_b=(\mathbf{p}_1,\mathbf{p}_2,\ldots)$ following the branching hierarchy constructed in \S\ref{sec:rib_gen}, yielding ordered root-to-tail spine paths. The complete spine structure is then represented as $\mathcal{S}=\{\mathbf{S}_1,\mathbf{S}_2,\ldots,\mathbf{S}_B\}$, where each branch spine $\mathbf{S}_b$ is an ordered sequence of spine points traversing one branch from root to tail (Fig.~\ref{fig:spine_label}).

\subsection{Skinning Weight Computation}
\label{sec:weights_gen}
To establish smooth correspondence between the mesh and the proposed rib-spine representation, we compute sparse skinning weights $\mathbf{W}=(\mathbf{W}^r,\mathbf{W}^s)$, where $\mathbf W^r=[\tilde w^{r}_{ik}]\in\mathbb R^{N\times K_r}$maps ribs to mesh vertices and $\mathbf W^s=[\tilde w^{s}_{ik}]\in\mathbb R^{N\times K_s}$ maps spine points to mesh vertices through nearest-point projection and adaptive Gaussian skinning weights \cite{singh1998wires}. The generated skinning weights $\mathbf{W}^r,\mathbf{W}^s$ support further rib-driven and spine-driven deformation and dynamics respectively.

\paragraph{Rib Skinning Weight.}
For each mesh vertex $\mathbf{v}_i$ and rib polyline
$R_k=(\mathbf r_k^1,\mathbf r_k^2,\ldots,\mathbf r_k^{M_k})$,
we first identify the nearest rib edge by minimizing the Euclidean point-to-edge distance
\begin{equation}
e_{ik}
=
\operatorname*{argmin}_{e_k^j=(\mathbf r_k^j,\mathbf r_k^{j+1})}
\|\mathbf v_i-\Pi_{e_k^j}(\mathbf v_i)\|_2,
\end{equation}
where $e_k^j$ denotes the line segment between two neighboring rib vertices and $\Pi_{e_k^j}(\mathbf v_i)$ denotes the orthogonal projection of $\mathbf v_i$ onto edge $e_k^j$. The corresponding Euclidean point-to-edge distance is then computed as $d_{ik}
= \|\mathbf v_i-\Pi_{e_{ik}}(\mathbf v_i)\|_2.$ Based on the projected distances $d_{ik}$, we compute adaptive Gaussian skinning weights
\begin{equation}
\tilde{w}^r_{ik}
=
\max\!\left(
e^{-d_{ik}^2/(2\sigma^2)}
-
w_{\min},
0
\right),
\label{eq:soft_weight}
\end{equation}
where $\sigma$ denotes the adaptive bandwidth and $w_{\min}$ is a small cutoff threshold. Row normalization produces the rib skinning matrix $\mathbf{W}^r$. To adapt the support range to local rib density, we further tie $\sigma$ to the inter-rib spacing $\Delta=L_p/K$ through $\sigma
=
\frac{n\Delta}
{\sqrt{-2\ln w_{\min}}}$,
so that each rib primarily supports a controlled number of neighboring ribs with spatially smooth decay.

\paragraph{Spine Skinning Weight.}
In addition to rib-based weights, we further compute a sparse spine skinning matrix $\mathbf{W}^s$ over the spine points. For each mesh vertex $\mathbf{v}_i$ and spine point $\mathbf{s}_k$, we compute the Euclidean point-to-point distance $d^{p}_{ik} = \|\mathbf v_i-\mathbf s_k\|_2.$ Based on these distances, we compute adaptive Gaussian skinning weights using the same Gaussian formulation as the rib-based weights, followed by row normalization. Spine points shared at Y-junctions therefore naturally share the same weight column, preserving branch consistency across the mesh. Vertices outside the effective support of all spine points are rigidly attached to their nearest spine point to avoid isolated unsupported regions.
\subsection{Deformation}
\label{sec:rib_deformation}
Given the extracted rib-spine representation $\mathcal{F}=(\mathcal{R},\mathcal{S},\mathbf{W})$ and the input mesh $\mathcal{M}=(\mathbf{V},\mathbf{T})$, we define a deformation operator
$\mathcal{M}'=\mathcal{D}(\mathcal{M},\mathcal{F})$
that generates a deformed mesh $\mathcal{M}'$. The operator $\mathcal{D}$ consists of six rib-driven deformation primitives, including uniform scaling, anisotropic scaling, translation, rotation, local deformation, and cross-section reshaping, as well as three spine-driven deformation primitives, including stretching, bending, and twisting. Rib-driven deformations first manipulate one or multiple ribs and propagate the induced displacements to the mesh through the skinning weights, followed by spine updates from the deformed mesh. Conversely, spine-driven deformations first deform the spine and reconstruct the mesh through preserved local coordinates, followed by rib updates. This bidirectional rib-mesh-spine coupling enables accurate and intuitive deformation control over general free-form geometries.

\paragraph{Rib-driven Deformations.}
In rib-driven deformations, for each mesh vertex $\mathbf{v}_i$, we precompute its closest projection point $\mathbf{p}_{ik}$ on rib $R_k$ (\S\ref{sec:weights_gen}). Given a deformed rib $R_k'$, the induced rib displacement is propagated to the mesh through the soft-cutoff Gaussian skinning weights $\tilde w^r_{ik}$:
\begin{equation}
\mathbf{v}_i'
=
\mathbf{v}_i
+
\tilde w^r_{ik}
\bigl(
\mathbf{p}_{ik}'-\mathbf{p}_{ik}
\bigr),
\label{eq:rib_lbs}
\end{equation}
where $\mathbf{p}_{ik}'$ denotes the deformed projection point corresponding to the same edge-parametric coordinate. More specifically, we define six rib-driven deformation primitives, including uniform scaling, anisotropic scaling, translation, rotation, local deformation, and cross-section reshaping. In uniform scaling, projection points are radially scaled around the rib centroid $\bar{\mathbf r}_k$: $\mathbf{p}_{ik}'-\mathbf{p}_{ik} = (s-1)(\mathbf{p}_{ik}-\bar{\mathbf r}_k),$ where $s\in\mathbb{R}^{+}$ denotes the uniform scaling factor. In anisotropic scaling, projection points are independently scaled along each axis using per-axis scaling factors $(s_x,s_y,s_z)$: $\mathbf{p}_{ik}'-\mathbf{p}_{ik} = (\operatorname{diag}(s_x,s_y,s_z)-\mathbf I)(\mathbf{p}_{ik}-\bar{\mathbf r}_k).$ In translation, a uniform displacement vector $\mathbf d\in\mathbb R^3$ is applied to the rib: $\mathbf{p}_{ik}'-\mathbf{p}_{ik}=\mathbf d.$ In rotation, the rib is rigidly rotated around its centroid $\bar{\mathbf r}_k$ with rotation matrix $\mathbf Q$: $\mathbf{p}_{ik}'-\mathbf{p}_{ik} = (\mathbf Q-\mathbf I)(\mathbf{p}_{ik}-\bar{\mathbf r}_k).$

For local deformation, a control projection point $\mathbf p_{ik_0}$ is displaced by $\mathbf d$, with the deformation smoothly decaying along the rib arc length:
\begin{equation}
\mathbf p_{ik}'
=
\mathbf p_{ik}
+
\mathbf d
\exp\!\left(
-\frac{(s_k-s_{k_0})^2}
{2\sigma_{\mathrm{drag}}^2}
\right),
\end{equation}
where $s_k$ denotes the cumulative arc length coordinate of the projection point along the rib and $\sigma_{\mathrm{drag}}$ controls the spatial support of the deformation. For cross-section reshaping, we operate in the rib-local coordinate system $(\mathbf e_u,\mathbf e_v,\mathbf n_k)$ centered at $\bar{\mathbf r}_k$. Each projection point is decomposed into in-plane coordinates and an out-of-plane offset along the plane normal $\mathbf n_k$. The in-plane coordinates are converted into polar coordinates $(\rho_{ik},\theta_{ik})$, where $\rho_{ik}$ denotes the radial distance from the rib center and $\theta_{ik}$ the angular direction. We then replace the original radius $\rho_{ik}$ with a target radius $\rho_{ik}^{\star}=\bar\rho\,\eta_\chi(\theta_{ik})$, where $\eta_\chi(\theta)$ denotes the radial profile of the target cross-section template $\chi$ and $\bar\rho$ is the mean rib radius. The reshaped projection point is reconstructed by replacing only the in-plane radius while preserving the angular direction and out-of-plane offset, and is finally blended with the original projection point to produce smooth cross-section deformation.

To further support both single-rib and multi-rib deformation, given a selected rib subset $\mathcal{K}\subseteq\{1,\ldots,K\}$, each rib independently produces a local displacement field through Eq.~\eqref{eq:rib_lbs}, and the resulting deformations are accumulated on the mesh: $\mathbf{v}_i' = \mathbf{v}_i + \sum_{k\in\mathcal{K}} \tilde w^r_{ik} \bigl( \mathbf{p}_{ik}'-\mathbf{p}_{ik} \bigr).$ Since the soft-cutoff Gaussian weights have compact support, the resulting deformations remain spatially localized even for disjoint rib selections.

\begin{figure}[t]
    \centering
    \includegraphics[width=\linewidth]{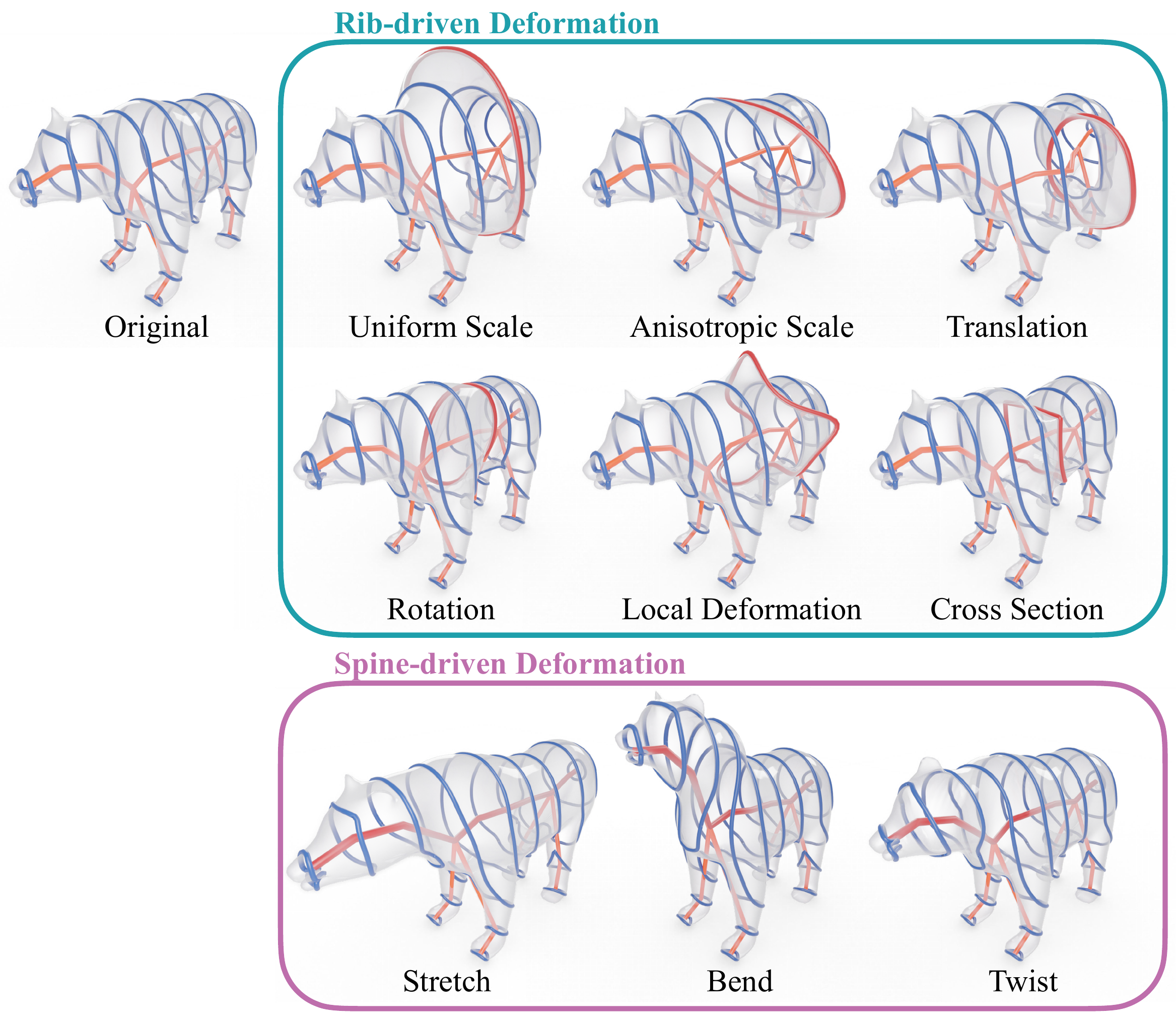}
    \caption{%
        \textbf{Deformation primitive gallery.}
        Gallery of \textbf{Fishbone} parametric deformations applied to the bear mesh.
        \emph{Top two rows} show six rib-driven variants produced by
        editing a single rib and propagating through
        projection-based skinning (Eq.~\eqref{eq:rib_lbs}): uniform 
        scaling, anisotropic scaling, translation, rotation, local
        deformation, and cross-section reshape toward a square target.
        \emph{Bottom row} shows the three spine-driven operations:
        stretching and bending deform the spine and propagate to the mesh,
        while twisting keeps the spine fixed and applies an in-plane
        rotation of each vertex's offset around the spine in the
        local coordinate frame. All nine variants are generated from a single input mesh by
        adjusting a small number of low-dimensional parameters.
    }
    \label{fig:deformation_gallery}
    \vspace{-15pt}
\end{figure}

\paragraph{Spine-Driven Deformations}
\label{sec:spine_deform}
In spine-driven deformations, the spine is first deformed from its initial configuration $\mathcal S=\{\mathbf p_k\}$ to a new configuration $\mathcal S'=\{\mathbf p_k'\}$, and the induced spine displacement is propagated to the mesh through the spine skinning weights $\tilde w^s_{ik}$:
\begin{equation}
\mathbf v_i'
=
\mathbf v_i
+
\sum_k
\tilde w^s_{ik}
\bigl(
\mathbf p_k'-\mathbf p_k
\bigr).
\label{eq:spine_lbs}
\end{equation}
We define three spine-driven deformation primitives: stretching, bending, and twisting. Let $L_0$ denote the total rest spine arc length and $(\mathbf{T}_0(t), \mathbf{N}_0(t), \mathbf{B}_0(t))$ the parallel-transport frame along the rest spine $\mathcal{S}_0$, where $\mathbf{T}_0(t)$ is the unit tangent and $\mathbf{N}_0(t), \mathbf{B}_0(t)$ are two orthonormal cross-sectional directions spanning the normal plane perpendicular to the spine.
In stretching, the spine is elongated or compressed along its arc length around an anchor point: $\ell_1(k) = a + s\bigl(\ell_0(k) - a\bigr), \; a = t_a L_0,$
where $\ell_0(k) \in [0, L_0]$ is the cumulative arc length of the $k$-th rest spine point measured from the spine root, $t_a \in [0, 1]$ is the user-specified anchor parameter, $a$ is its arc-length position, and $s > 0$ is the stretching factor ($s = 1$ identity, $s > 1$ elongates, $s < 1$ compresses). The deformed spine $\mathcal{S}_1$ is then propagated to the mesh through Eq.~\eqref{eq:spine_lbs}. In bending, the tail segment $\{t > t_a\}$ of the spine is rigidly rotated around the anchor point $\mathcal{S}(t_a)$ about a bending axis $\mathbf{a} \in \{\mathbf{N}_0(t_a), \mathbf{B}_0(t_a)\}$, while the remaining portion stays fixed. The deformed spine is subsequently propagated to the mesh through Eq.~\eqref{eq:spine_lbs}. In twisting, the spine geometry itself remains fixed while the local vertex offsets are rotated around the spine in the parallel-transport frame $(\mathbf{N}_0, \mathbf{B}_0)$. The torsion angle $\psi(t)$ increases progressively from $0$ to $\psi_{\max}$ over the user-specified interval $[t_{\mathrm{start}}, t_{\mathrm{end}}]$ and remains constant outside this range. The local cross-sectional coordinates $(u_i, v_i)$ are rotated in the normal plane by $\psi(t_i)$, producing the rotated offsets $(u_i', v_i')$. The twisted mesh vertex is then reconstructed as $\mathbf{v}_i' = \mathcal{S}_0(t_i) + u_i' \mathbf{N}_0(t_i) + v_i' \mathbf{B}_0(t_i).$ For extremely large spine-driven deformations, remeshing can better resolve distorted regions.

\paragraph{Composition and Bi-directional Coupling}
Rib- and spine-driven operations can be chained arbitrarily through the shared mesh representation. After a rib-driven edit deforms the mesh, the spine is updated by aggregating nearby mesh displacements through the precomputed spine skinning weights, yielding naturally smooth spine motion. Conversely, after a spine-driven edit updates the mesh, the rib polylines are updated together with the deformed surface, preserving the coupled rib-spine structure without re-extraction.

\section{Fishbone Dynamics and Animation}
\label{sec:dynamics}

\begin{figure}[h]
    \centering
    \includegraphics[width=\linewidth]{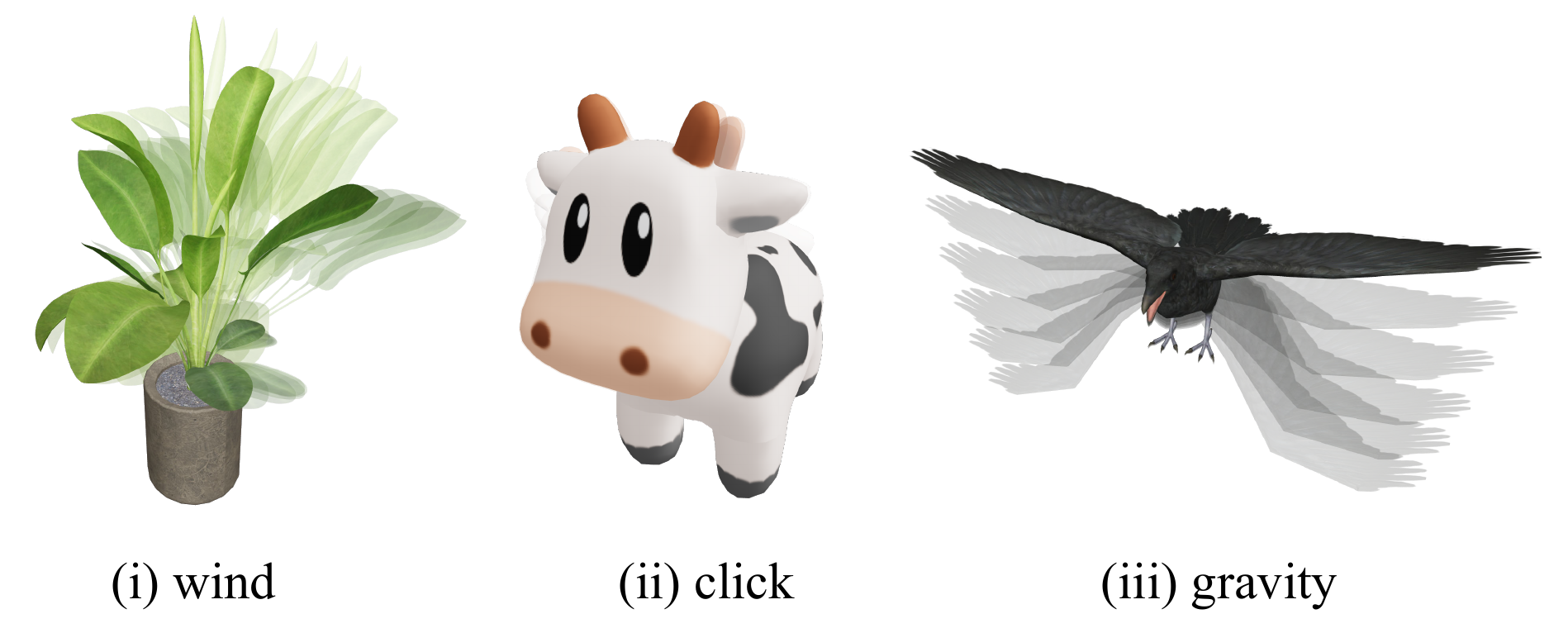}
    \caption{%
        \textbf{Reduced simulation.} 
        We demonstrate visually plausible and interactive reduced-space dynamics with the proposed method: \emph{(i) Plant sway under wind:} branch-aware stretch and length-normalized bending, combined with root-pinned key points and a tip-biased wind ramp, produce coherent plant motion. The drag-form wind model adds turbulence and tangent-perpendicular projection to prevent stems from sliding along their own axes, while each branch is simulated independently to avoid spurious cross-branch coupling. \emph{(ii) Click-to-jiggle:} user impulses applied on mesh vertices are projected to nearby spine key points with Gaussian falloff and dissipate through stretch, bending, and Rayleigh damping. \emph{(iii) Gravity-pulled wings:} anchored bird wings deform downward under gravity until balanced by bending resistance, with the resulting droop controlled by the bending stiffness.
    }
    \label{fig:dyn_demos}
    \vspace{-10pt}
\end{figure}

The proposed rib-spine representation further supports reduced-space dynamics by treating the spine as low-dimensional simulation coordinates lifted to the mesh through $\mathbf{W}^s$, as well as lightweight keyframe animation through interpolation between edited \textbf{Fishbone} states. We treat the automatically extracted spine $\mathcal{S}$ as the reduced coordinate, equip it with lumped mass, damping, and a lightweight elastic potential, and lift the simulated motion back to the mesh through the precomputed spine weights $\mathbf{W}^s$. Semi-implicit Euler then produces real-time physically inspired secondary motion, such as plant sway and click-to-jiggle response, without simulating every mesh vertex. The transpose $(\mathbf{W}^s)^\top$ also projects user interaction and external forces (e.g., wind and gravity) back to reduced space, enabling interactive effects such as plant sway, click-to-jiggle response, and gravity-driven wing droop (Fig.~\ref{fig:dyn_demos}).

\subsection{Reduced State and Equations of Motion}
\label{sec:dyn_state}

The dynamic state is the time-dependent spine, lifted to the mesh through the spine skinning matrix:
\begin{subequations}
\label{eq:dyn_kinematics}
\begin{align}
    q^{t} &= \mathcal{S}^{t} = (\mathbf{p}_1^{t}, \ldots, \mathbf{p}_K^{t}), \label{eq:dyn_state} \\
    \mathbf{V}^{t} &= \mathbf{V}^0 + \mathbf{W}^s \, \bigl(\mathcal{S}^{t} - \mathcal{S}^0\bigr), \label{eq:dyn_lift}
\end{align}
\end{subequations}
where $q^t$ is the reduced dynamic state stacking the $K$ time-dependent spine key points $\mathbf{p}_i^t$, with $\mathbf{V}^0$ and $\mathcal{S}^0$ the rest mesh and rest spine. Eq.~\ref{eq:dyn_lift} uses the same linear-blend translation operator~\cite{magnenat1989joint} as the static spine-driven deformation (Eq.~\ref{eq:spine_lbs}), unifying static editing and dynamic simulation under a single mesh-lift pass. The dynamics follow the reduced-coordinate Lagrangian~\cite{barbivc2005real} with mass matrix $\mathbf{M}_q$, damping $\mathbf{D}_q$, elastic potential $E(q)$, and total external force $\mathbf{F}_q^{\text{ext}}$ in the reduced spine coordinates,
\begin{equation}
    \mathbf{M}_q \ddot q + \mathbf{D}_q \dot q + \nabla_q E(q) = \mathbf{F}_q^{\text{ext}},
    \label{eq:dyn_eom}
\end{equation}
where $\mathbf{F}_q^{\text{ext}} \in \mathbb{R}^{K \times 3}$ aggregates spine-direct excitations (e.g., wind, impulse) together with the pull-back of any mesh-side per-vertex force matrix $\mathbf{F}_V \in \mathbb{R}^{N \times 3}$ through the adjoint of the lift, $(\mathbf{W}^s)^\top \mathbf{F}_V$, where $\mathbf{W}^s \in \mathbb{R}^{N \times K}$ is the spine skinning matrix. 

We approximate $\mathbf{M}_q$ as a diagonal lumped-mass matrix induced by the surface area associated with each spine key point. Following the barycentric lumped-mass discretization~\cite{meyer2003discrete}, each triangle contributes one third of its area to its incident vertices, whose masses are projected to the reduced spine coordinates through the same skinning matrix $\mathbf{W}^s$ used for mesh lifting and unit-mean normalized to obtain the per-key-point spine masses:
\begin{equation}
\tilde m_i = \sum_{j=1}^{N} (\mathbf{W}^s)_{ji}
\left(\rho \sum_{f \in \mathcal{N}(j)} \tfrac{A_f}{3}\right),
\qquad
m_i = \frac{\tilde m_i}{\tfrac{1}{K}\sum_{r=1}^{K} \tilde m_r + \varepsilon},
\label{eq:dyn_mass}
\end{equation}
where $A_f$ denotes the area of triangle $f$,  $\rho$ is a per-part surface density, $\mathcal{N}(j)$ denotes the set of triangles incident to vertex $j$, and the reduced mass matrix is $\mathbf{M}_q = \operatorname{diag}(m_1 \mathbf{I}_3, \ldots, m_K \mathbf{I}_3)$.
Area weighting yields inertia proportional to surface area for thin-shell and leaf-like geometries, while unit-mean normalization keeps stiffness, damping, and wind parameters approximately resolution independent.

As for damping, rather than using a single uniform coefficient, we adopt a Rayleigh-style decomposition~\cite{rayleigh1896theory} with independently tunable global, stretch, and bending modes,
$\mathbf{D}_q \dot{\mathbf q} = \alpha \mathbf{M}_q \dot{\mathbf q} + \beta_e \mathbf{D}_e \dot{\mathbf q} + \beta_b \mathbf{D}_b \dot{\mathbf q}$, where $\alpha \mathbf{M}_q$ damps low-frequency global motion, $\mathbf{D}_e$ dissipates only the stretch mode through along-edge relative velocities, and $\mathbf{D}_b$ damps only the curvature mode via the second-order stencil $\dot{\mathbf{p}}_{i-1} - 2\dot{\mathbf{p}}_i + \dot{\mathbf{p}}_{i+1}$. Both operators are momentum preserving through equal-and-opposite per-element contributions, and $\beta_e=\beta_b=0$ reduces to uniform damping. We then integrate Eq.~\ref{eq:dyn_eom} using semi-implicit Euler~\cite{hairer2006geometric}. At each substep, the spine velocities and positions are updated as
\begin{equation}
\label{eq:dyn_integrator}
\dot{\mathbf{p}}_i^{n+1}
=
\dot{\mathbf{p}}_i^{n}
+
\Delta t \, \frac{\mathbf{f}_i^{n}}{m_i},
\mathbf{p}_i^{n+1}
=
\mathbf{p}_i^{n}
+
\Delta t \, \dot{\mathbf{p}}_i^{n+1}.
\end{equation}
After this update, we apply a hard-pin projection enforcing $\dot{\mathbf{p}}_i = \mathbf{0}$ and $\mathbf{p}_i = \mathbf{p}_i^{0}$ for all $i \in \mathcal{P}$.
The force $\mathbf{f}_i^{n}$ combines elastic, bending, damping, and external terms.

\subsection{Elastic Potential}
\label{sec:dyn_energy}

The elastic potential decomposes into an axial stretch term and a discrete-rod bending term, $ E(q) = E_{\text{stretch}}(\mathcal{S}) + E_{\text{bend}}(\mathcal{S}),$ each parameterized by a single scalar stiffness.

For the stretch energy, let $\ell_i^0 = \|\mathbf{p}_{i+1}^0 - \mathbf{p}_i^0\|$ denote the rest length of spine edge $(i,i+1)$, and let $\bar{\ell}^0$ denote the mean rest length over all spine edges within the part. The rest-length-scaled axial-spring energy
\begin{equation}
    E_{\text{stretch}} = \tfrac{1}{2} \sum_{i=1}^{K-1} k_s \, \sigma_i \bigl(\|\mathbf{p}_{i+1} - \mathbf{p}_i\| - \ell_i^0\bigr)^2,
    \qquad \sigma_i = \frac{\bar\ell^0}{\ell_i^0 + \varepsilon},
    \label{eq:dyn_stretch}
\end{equation} where $k_s > 0$ is a user-controlled stretch stiffness and $\sigma_i$ normalizes the response with respect to the rest edge length, making stiffness invariant to spine resampling while preserving the mean-edge response defined by $k_s$. In terms of discrete-rod bending~, let $\mathbf{t}_i = (\mathbf{p}_{i+1} - \mathbf{p}_i) / \|\mathbf{p}_{i+1} - \mathbf{p}_i\|$ be the unit tangent of edge $i$.
At each interior key point we form the length-normalized discrete curvature $\boldsymbol{\kappa}_i$ (with rest value $\boldsymbol{\kappa}_i^0$) and accumulate the per-triple bending energy~\cite{bergou2008discrete}:
\begin{subequations}
\label{eq:dyn_bending_full}
\begin{align}
    \boldsymbol{\kappa}_i &= \frac{2 \, (\mathbf{t}_i - \mathbf{t}_{i-1})}{\ell_{i-1}^0 + \ell_i^0}, \label{eq:dyn_curvature} \\
    E_{\text{bend}}       &= \tfrac{1}{2} \sum_{i=2}^{K-1} k_b \, \bar\ell_i^0 \, \bigl\|\boldsymbol{\kappa}_i - \boldsymbol{\kappa}_i^0\bigr\|^2,
    \qquad \bar\ell_i^0 = \tfrac{1}{2}(\ell_{i-1}^0 + \ell_i^0), \label{eq:dyn_bend}
\end{align}
\end{subequations} where $k_b > 0$ is a user-controlled bending stiffness and $\bar\ell_i^0$ normalizes the curvature response with respect to local edge length, making the bending behavior invariant to spine resampling. This yields stiffness approximately invariant to spine resampling, avoiding the $1/\ell^3$ scaling drift of the simpler rest-Laplacian energy $\tfrac{1}{2} k_b \|\mathbf{p}_{i-1} - 2\mathbf{p}_i + \mathbf{p}_{i+1} - \Delta_i^0\|^2$, which we retain only as a legacy fallback. For branched spines, each branch is simulated independently: stretch energies are accumulated over within-branch edges with shared junction edges counted only once, while bending energies are evaluated separately along each branch path to avoid spurious cross-branch coupling. In addition, a subset $\mathcal{P}$ of spine key points is constrained to remain at the rest configuration through the Dirichlet condition $\mathbf{p}_i^{t} = \mathbf{p}_i^{0}, \forall\, i \in \mathcal{P},$ enforced after each integration substep by setting $\dot{\mathbf{p}}_i=\mathbf{0}$ and projecting $\mathbf{p}_i$ back to $\mathbf{p}_i^{0}$ for all $i \in \mathcal{P}$.

\subsection{External Forces}
\label{sec:dyn_forces}
External excitation enters either as direct reduced-space forces on the spine or as mesh-side per-vertex forces projected onto the reduced coordinates through $\mathbf W^s$. We support wind, gravity, user interaction impulses, and general mesh-side forces. For wind~\cite{wejchert1991animation}, we model coherent sway through a sinusoidal driving term together with aerodynamic damping induced by the relative velocity between the ambient flow and the moving spine:
\begin{equation}
\label{eq:dyn_wind_full}
\mathbf{F}_i^{\mathrm{wind}}=
a_i A \hat{\mathbf w}\sin(\omega t+i\Delta\phi),
\mathbf{F}_i^{\mathrm{drag}}=
c_d a_i\bigl(\mathbf I-\mathbf t_i\mathbf t_i^\top\bigr)\bigl(\mathbf u_{\mathrm{wind}}(\mathbf p_i,t)-\dot{\mathbf p}_i\bigr).
\end{equation}
Here $\hat{\mathbf w}$ denotes the wind direction, $A$ the wind amplitude, $\omega$ the temporal frequency, and $\mathbf t_i$ the local spine tangent. The spatial ramp $
a_i=(d_i/d_{\max})^p
$
scales the excitation according to the distance from the nearest pinned root, producing larger oscillations toward free ends. We also support gravity through $\mathbf F_i^{\mathrm{grav}}=m_i\mathbf g$, where the spine mass $m_i$ is aggregated from the area-weighted mesh mass. User interaction is modeled as a localized impulse. Given a click position $\mathbf{x}_c$ and impulse vector $\mathbf{J}$, we distribute the impulse to nearby spine key points using Gaussian weights centered at $\mathbf{x}_c$
\begin{equation}
\omega_i^{\text{imp}}
=
\exp\!\left(
-\frac{\|\mathbf{p}_i-\mathbf{x}_c\|^2}{2\sigma_{\text{imp}}^2}
\right),
\dot{\mathbf{p}}_i
\leftarrow
\dot{\mathbf{p}}_i
+
\omega_i^{\text{imp}}
\frac{\mathbf{J}}{m_i},
\label{eq:dyn_impulse}
\end{equation}
where $\sigma_{\text{imp}}$ controls the spatial support. Setting $\sigma_{\text{imp}}\to0$ applies the impulse only to the nearest key point. Moreover, general external forces defined on the mesh surface, such as mesh-side contact, drag, ambient fields sampled at vertices, and user interactions, are projected to the reduced spine state through $(\mathbf{W}^s)^\top$, the adjoint of the spine-to-mesh lift in Eq.~\ref{eq:dyn_lift}. The same $\mathbf{W}^s$ therefore acts as both lifting and pull-back operator, ensuring consistent bidirectional coupling without additional calibration.

\subsection{Mesh Lifting}
\label{sec:dyn_lifting}

For reduced-space dynamics, we further explore a cylindrical lifting scheme based on the spine's parallel-transport frame to better preserve rotational deformation for slender structures such as plants or tails. At rest, each mesh vertex $\mathbf v_i$ is associated with its nearest spine segment and encoded by local cylindrical coordinates $(\ell_i,\alpha_i,u_i,v_i)$, where $\ell_i\in[0,1]$ denotes the parametric position along the segment, $\alpha_i$ the displacement along the local tangent direction, and $(u_i,v_i)$ the offsets in the local normal-binormal plane, together with its perpendicular distance $d_i$ to the spine. Let $\mathbf v_i^{\mathrm{disp}}=\mathbf v_i^0+\sum_{k=1}^{K}(\mathbf{W}^s)_{ik}(\mathbf p_k^t-\mathbf p_k^0)$ denote the per-vertex displacement lift induced by Eq.~\ref{eq:dyn_lift}. During simulation, the local frame is updated on the deformed spine and the final lifted vertex is computed as
\begin{equation}
\label{eq:dyn_cyl}
\mathbf v_i^t=\mathbf v_i^{\mathrm{disp}}+\lambda_i\bigl(\tilde{\mathbf v}_i^t-\mathbf v_i^{\mathrm{disp}}\bigr), \lambda_i=\exp(-d_i^2/(2\sigma_s^2)).
\end{equation}
where $\tilde{\mathbf v}_i^t$ is the cylindrical reconstruction in the deformed local frame and $\sigma_s$ controls the spatial blending range. Vertices near the spine inherit local rotational motion, while distant regions smoothly reduce to the displacement-based lift.

\begin{figure}[t]
    \centering
    \includegraphics[width=\linewidth]{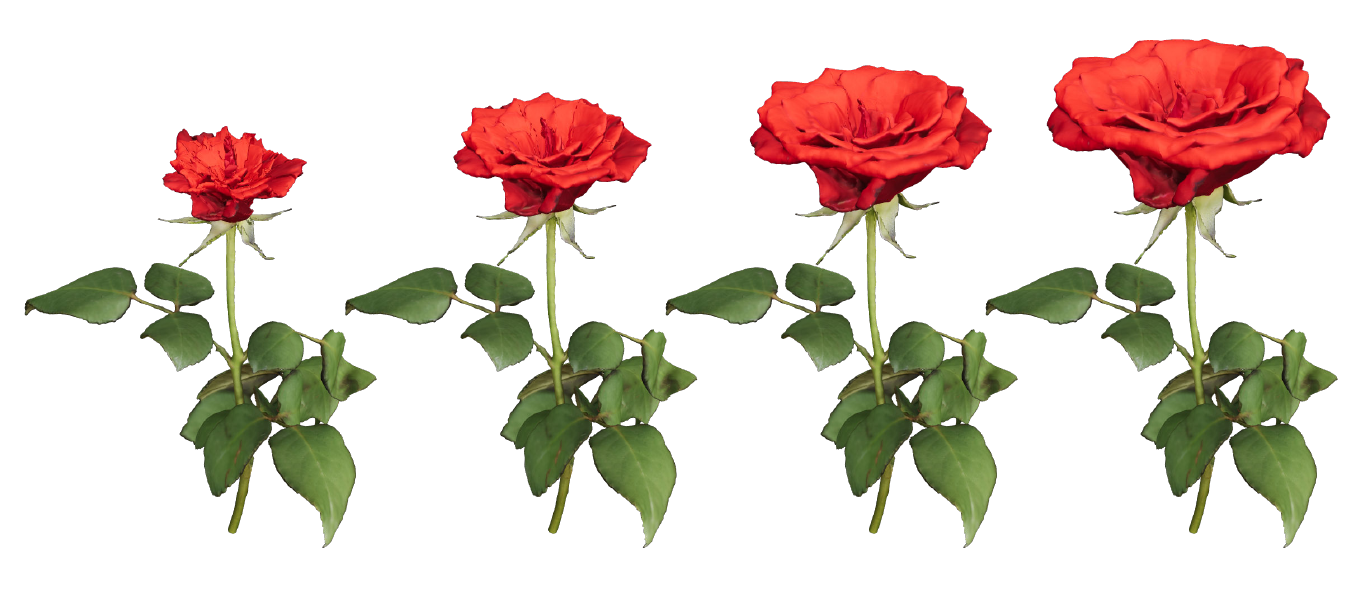}
    \caption{\textbf{Keyframe animation.} A rose blooming from bud to full bloom, authored as a few \textbf{Fishbone} edits and replayed by the per-vertex linear interpolation between keyframes (\S\ref{sec:animation}).}
    \label{fig:animation}
    \vspace{-10pt}
\end{figure}

\subsection{Keyframe Animation}
\label{sec:animation}

The proposed rib-spine representation also supports keyframe animation. Users can author a sparse set of target poses through the rib- and spine-driven deformation primitives and store each edited state as a keyframe containing the deformed mesh, rib polylines, and spine key points. Given $M$ ordered keyframes $\{\mathbf q_1,\ldots,\mathbf q_M\}$, we generate in-between motions through
$\mathbf q = (1-\tau)\mathbf q_m + \tau \mathbf q_{m+1}, \tau\in[0,1],$ where $\tau$ denotes the interpolation parameter between keyframes, and the interpolation is applied independently to mesh vertices, rib polylines, and spine key points. The proposed keyframe animation pipeline is particularly suitable for replaying authored target motions. Fig.~\ref{fig:animation} shows a rose-blooming sequence generated from a sparse set of keyframes with the proposed \textbf{Fishbone} representation.

\begin{figure*}[!h]
    \centering
    \includegraphics[width=\textwidth, height=0.82\textheight, keepaspectratio]{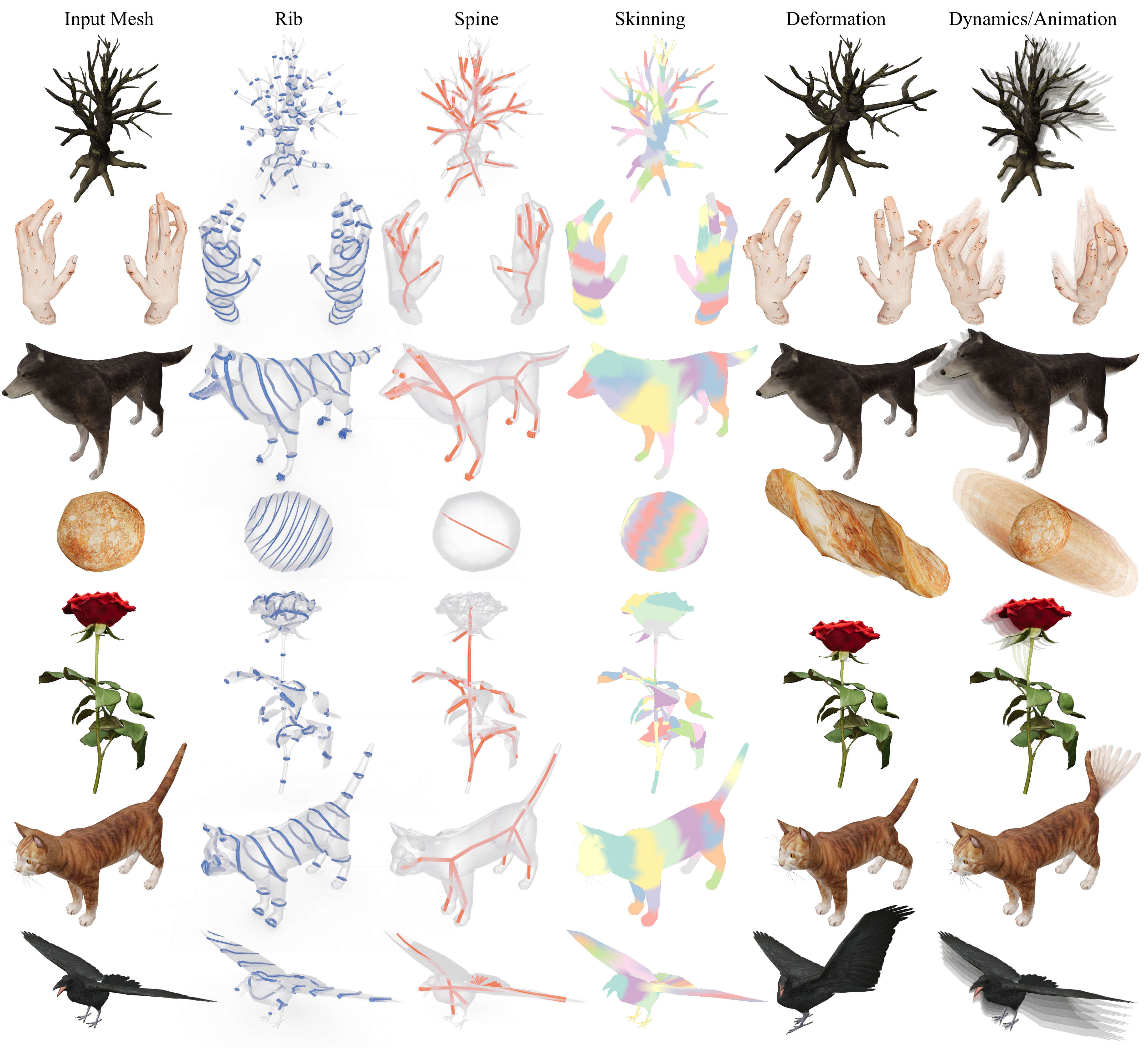}
    \caption{
        \textbf{Gallery.}
        The proposed method generates stable and accurate rib-spine structures together with the corresponding skinning weights for diverse mesh inputs, without requiring per-asset tuning, category-specific templates, or manual rigging. Based on the generated representation, our framework further supports real-time, geometrically consistent deformation and efficient reduced-space dynamics simulation.
        }

    \label{fig:more_demo}
\end{figure*}

\section{Evaluation}
\label{sec:evaluation}
In this section, we report implementation details, evaluate the runtime performance of the proposed framework, and assess the key components of our pipeline through extensive ablation studies, including rib extraction, spine construction, branching handling, and reduced-space deformation lifting.

\begin{table}[!h]
    \centering
    \small
    \begin{tabular*}{\linewidth}{@{\extracolsep{\fill}}lrrrrr@{}}
        \toprule
        \textbf{Stage} & $N$ & \textbf{mean$\pm$std} & \textbf{median} & \textbf{min} & \textbf{max} \\
        \midrule
        \multicolumn{6}{@{}l}{\textit{Preprocessing (offline, one-time per asset, seconds)}} \\
        Rib extraction              & $2{,}565$ & $50.16 \pm 53.88$ & $34.80$ & $0.41$ & $663.49$ \\
        Spine construction          & $2{,}565$ &  $4.01 \pm  2.28$ &  $3.61$ & $0.41$ &  $21.05$ \\
        Weight computation          & $2{,}565$ & $30.30 \pm 40.62$ & $17.42$ & $0.02$ & $568.72$ \\
        Total preprocessing         & $2{,}565$ & $84.48 \pm 92.78$ & $58.04$ & $0.92$ & $1{,}143.20$ \\
        \midrule
        \multicolumn{6}{@{}l}{\textit{Deformation (online, per edit, milliseconds)}} \\
        Uniform scaling      & $37$ & $19.39 \pm 55.55$ &  $1.34$ & $0.37$ & $247.94$ \\
        Anisotropic scaling  & $37$ & $20.30 \pm 58.66$ &  $1.18$ & $0.38$ & $264.26$ \\
        Translation          & $37$ &  $8.75 \pm 22.45$ &  $0.78$ & $0.29$ &  $98.62$ \\
        Rotation             & $37$ & $25.60 \pm 76.66$ &  $1.19$ & $0.42$ & $362.29$ \\
        Shape blend          & $37$ & $20.96 \pm 59.92$ &  $1.27$ & $0.46$ & $275.09$ \\
        Stretching           & $37$ &  $7.80 \pm 18.71$ &  $0.84$ & $0.35$ &  $84.43$ \\
        Bending              & $37$ &  $8.54 \pm 18.70$ &  $1.32$ & $0.68$ &  $87.32$ \\
        Twisting             & $37$ & $352 \pm 1{,}114$ & $12.74$ & $4.70$ & $5{,}240$ \\
        \midrule
        \multicolumn{6}{@{}l}{\textit{Dynamics (online, per simulation step, milliseconds)}} \\
        Physics only                &  $188$    &  $1.50 \pm  1.21$ &  $1.08$ & $0.20$ &   $8.00$ \\
        $+$ displacement lift        &  $188$    &  $4.88 \pm  5.79$ &  $3.00$ & $0.27$ &  $38.88$ \\
        $+$ cylindrical lift        &  $188$    & $39.49 \pm 56.44$ & $25.72$ & $1.40$ & $453.60$ \\
        \bottomrule
    \end{tabular*}
    \caption{\textbf{Runtime summary.} Wall-clock runtime across the offline preprocessing pipeline and the two online pipelines (deformation and dynamics) on the workstation above. $N$ is the number of meshes the stage was measured on, and \textbf{mean$\pm$std}, \textbf{median}, \textbf{min}, and \textbf{max} are computed across that run. The median is the typical cost, while the standard deviation and max capture the long upper tail driven by very large or non-manifold inputs. Preprocessing is measured per mesh on our $2{,}565$-mesh extraction run, in seconds. Deformation is measured per edit on a $37$-mesh subset spanning up to $1{,}690{,}154$ vertices, in milliseconds (one row per primitive of \S\ref{sec:rib_deformation}). Dynamics is measured per simulation step on the $188$-mesh dynamics benchmark, in milliseconds. The weight matrix is cached on disk after the first computation, so subsequent loads are millisecond-scale.}
    \label{tab:runtime_summary}
    \vspace{-10pt}
\end{table}

\subsection{Implementation Details}
\label{sec:impl}
Our pipeline is implemented in Python with GPU acceleration through \texttt{PyTorch}. The interactive interface supports both desktop and browser based editing, and the same backend is reused by the language model agent in \S\ref{sec:app_agent}. Rib counts, influence bandwidths, and spine scoring parameters are automatically derived per part rather than fixed globally. The complete list of constants and parameters is provided in Appendix~\ref{sec:default_params}.

\subsection{Results}

We demonstrate the effectiveness and expressiveness of the proposed framework through rib-spine extraction, skinning, controllable mesh deformation, and reduced-space dynamics results on the part-level Hunyuan3D~\cite{zhao2025hunyuan3d} dataset (Fig.~\ref{fig:more_demo}). The proposed method consistently extracts semantically meaningful and geometrically consistent control structures, enabling accurate and stable deformation, simulation, and animation across diverse generated shapes. In addition, we apply the proposed method to the Hunyuan3D~\cite{zhao2025hunyuan3d} part-level mesh dataset to jointly generate part meshes, rib-spine structures, and skinning weights, contributing the \textbf{Fishbone-136K} dataset. The dataset enables parametric mesh deformation, and reduced dynamics via the proposed rib-spine control representation.

\subsection{Performance}
\label{sec:performance}
We report the runtime performance of \textbf{Fishbone} across both the offline preprocessing pipeline for rib-spine representation generation and real-time deformation, simulation, and animation. All experiments are conducted on a workstation with an AMD EPYC 9354 32 core CPU, 1.5~TB RAM, and eight NVIDIA H100 NVL GPUs. Tab.~\ref{tab:runtime_summary} summarizes the per stage runtime, while the remainder of this section further breaks down the cost of each component.

\begin{figure}[!t]
    \centering
    \includegraphics[width=0.85\linewidth]{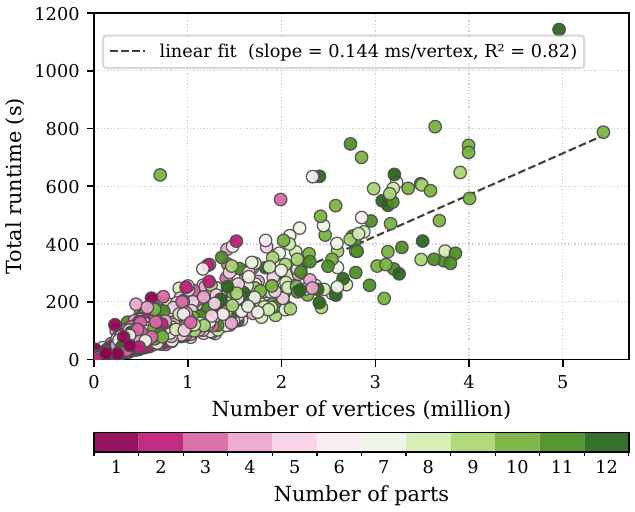}
    \caption{\textbf{Preprocessing runtime vs.\ mesh resolution.} Total per-mesh wall-clock runtime as a function of cleaned vertex count, with each point colored by the number of parts in the mesh ($1$-$12$). Runtime grows roughly linearly with mesh resolution, dominated by per-vertex weight queries against the rib polylines.}
    \label{fig:runtime_vs_vertices}
    \vspace{-10pt}
\end{figure}

\paragraph{Mesh-to-rib-spine extraction.}
The full preprocessing pipeline takes a median of $58.0$~s per mesh ($84.5$~s mean) across our $2{,}565$ meshes, with a long upper tail caused by oversize and non-manifold inputs (Tab.~\ref{tab:runtime_summary}). The preprocessing pipeline includes optional watertightness repair (\S\ref{sec:preprocess}), the heat method, rib extraction, spine construction, and skinning weight computation. Rib extraction dominates the runtime (median $34.8$~s), followed by weight computation ($17.4$~s), while spine construction is comparatively lightweight ($3.6$~s). As shown in Fig.~\ref{fig:runtime_vs_vertices}, runtime scales roughly linearly with cleaned vertex count, since the dominant weight queries scale with mesh resolution.

\paragraph{Deformation.}
Each rib- or spine-driven edit (\S\ref{sec:rib_deformation}) reduces to a sparse matrix-vector product against the cached $\mathbf{W}$. Across $37$ meshes spanning up to $1{,}690{,}154$ vertices, seven of the eight deformation primitives achieve a median runtime of approximately $1$~ms per edit, while twisting requires $12.7$~ms median (Tab.~\ref{tab:runtime_summary}), enabling stable real-time interactive editing.

\begin{figure}[!t]
    \centering
    \includegraphics[width=\linewidth]{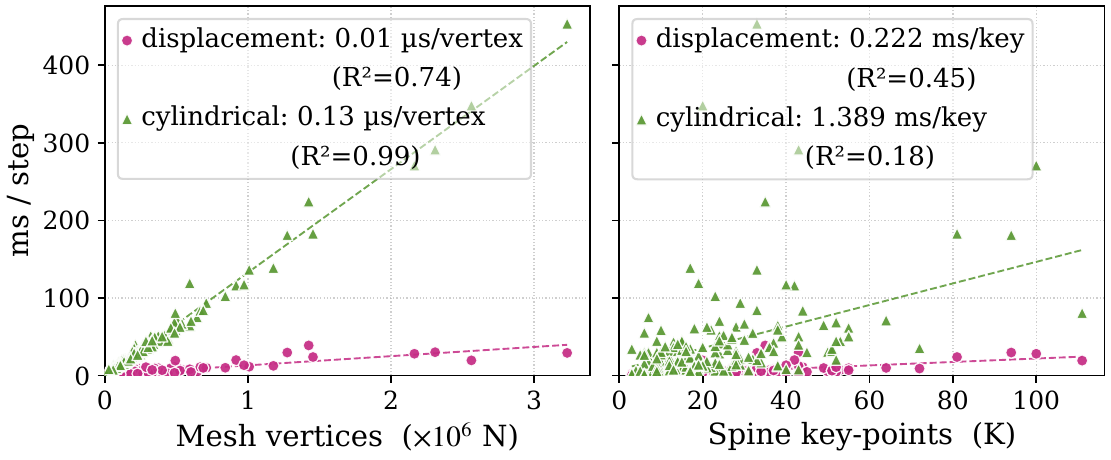}
    \caption{\textbf{Dynamics latency scaling.} Per-step latency scaling of the reduced-dynamics pipeline. \emph{Left:} latency vs.\ mesh vertex count $N$, with linear fits reported in $\mu$s/vertex. The cylindrical lift's per-vertex slope is roughly an order of magnitude steeper than the displacement lift's because each vertex requires walking the parallel-transport frame and evaluating its rest cylindrical coordinates. \emph{Right:} latency vs.\ total spine key-point count $K$, with linear fits in ms/key. The physics step itself is independent of mesh size and scales only with $K$. Displacement (circles) and cylindrical (triangles) lifts are fit independently in each panel. Legends report each fit's slope and its coefficient of determination $R^2 \in [0, 1]$, the fraction of latency variance the linear model explains ($R^2 \to 1$ confirms the scaling is genuinely linear, not just visually plausible).}
    \label{fig:dyn_ms_scaling}
    \vspace{-10pt}
\end{figure}

\paragraph{Reduced dynamics.}
We benchmark the integrator and the two mesh-lift modes of \S\ref{sec:dyn_lifting} across $188$ meshes spanning $K = 3$-$111$ spine points, $N=6.6$K-$3.2$M vertices, and $B=1$-$23$ spine branches. The semi-implicit Euler integrator runs at $1.1$~ms median ($1.5$~ms mean) per substep. Mesh lifting adds $1.9$~ms median for the displacement lift ($3.0$~ms total per frame) and $24.6$~ms median for the cylindrical lift ($25.7$~ms total per frame) on CPU (Tab.~\ref{tab:runtime_summary}). Per-step latency scales roughly linearly with both mesh vertex count and spine key-point count (Fig.~\ref{fig:dyn_ms_scaling}), while the cylindrical lift is more expensive due to the parallel-transport frame evaluation required per vertex. The reduction is substantial: instead of simulating all mesh vertices, the reduced dynamics operates on only $K \in [5, 2{,}058]$ spine key points, while the corresponding meshes contain $N \in [254, 5{,}432{,}594]$ vertices. Across the benchmark, this yields a median $K/N$ reduction ratio of only $0.007\%$ (mean: $0.064\%$, range: $0.001\%$-$24.36\%$).

\begin{figure*}[t]
    \centering
    \includegraphics[width=\linewidth]{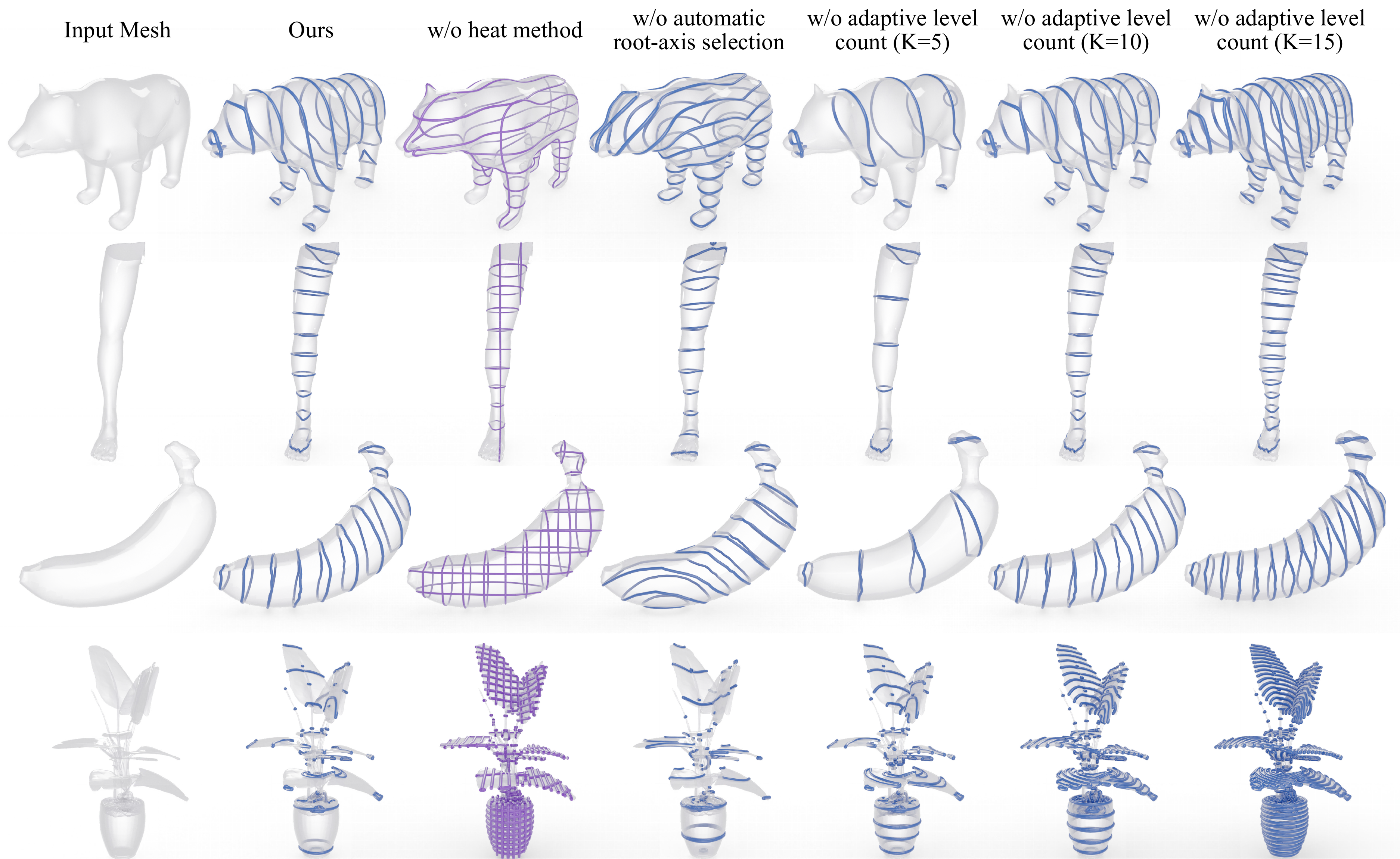}
    \caption{%
        \textbf{Rib extraction design choices.}
        Seven panels evaluate each component of our rib extraction on the same input mesh.
        \emph{Input mesh} is the cleaned source.
        \emph{Ours} runs the full pipeline: heat-method geodesic ribs with automatic root-axis selection and adaptive level count $K = \operatorname{clip}(\operatorname{round}(10\,L_p/L_o),\,3,\,10)$.
        The remaining five panels each remove one component:
        \emph{w/o heat method} replaces the geodesic iso-contours with a fixed stack of axis-aligned planes (\emph{grid ribs}),
        \emph{w/o automatic root-axis selection} pins the root face to a global axis instead of the per-part longest-extent axis $a^\star$,
        and the three \emph{w/o adaptive level count} panels run the full pipeline but with a hand-fixed $K \in \{5, 10, 15\}$ for every part.
        The first two ablations (\emph{w/o heat method} and \emph{w/o automatic root-axis selection}) both fail to slice the part along its natural direction: the grid baseline replaces geodesic iso-contours with fixed axis-aligned planes, while the wrong-axis variant runs the heat method but propagates from the wrong end, so the rib stack ends up misaligned with the object's principal axis.
        The fixed-level-count ablation instead surfaces an over- and under-sampling problem that is most visible on multi-part objects (e.g., the plant in row 4): $K=5$ under-resolves large parts (the stem), while $K=10$ and $K=15$ over-resolve small auxiliary parts (leaves, buds).
    }
    \label{fig:ablation_rib}
\end{figure*}

\subsection{Ablation Study}
\label{sec:ablation}
In this section, we assess the essential components of our proposed system, including rib extraction design choices, spine extraction scores, branching-tree handling, and mesh lifting strategies.

\paragraph{(a) Rib extraction design choices.} 
We ablate three components of rib extraction (\S\ref{sec:rib_gen}) that help ribs conform to local geometry while remaining coherent across general shapes: the heat-method geodesic field, automatic root-axis selection, and adaptive level count.

\emph{Adaptive heat method.}
We first evaluate the proposed adaptive heat-based rib extraction, which produces ribs aligned with local geometry. As shown in Fig.~\ref{fig:ablation_rib}, replacing geodesic iso-contours with fixed axis-aligned slicing planes leads to fragmented and geometrically inconsistent cross-sections on bent or branching structures. In contrast, our heat-based ribs follow the geodesic field and naturally conform to the underlying shape structure, particularly around limbs and thin protrusions. This highlights the importance of the adaptive heat method in preserving local geometric coherence across diverse shapes.

\emph{Automatic root-axis selection.}
Automatic root-axis selection ensures that the geodesic field propagates along the main extent of each part. Without it, fixing the root to a global axis (e.g., always min-$y$) misroots tilted or sideways-oriented structures, causing iso-contours to run lengthwise instead of forming meaningful cross-sections. As shown in Fig.~\ref{fig:ablation_rib}, this collapses the rib layout and loses the underlying structural organization.

\emph{Adaptive level count.}
Using a fixed rib count for all parts produces inconsistent sampling across meshes with different scales and shapes. As shown in Fig.~\ref{fig:ablation_rib}, small rib counts under-sample large structures, while large rib counts over-sample small auxiliary parts with redundant near-coincident loops. Adapting the rib count to the relative part extent yields more consistent rib spacing and structural coverage across diverse parts.

\begin{figure}[h]
    \centering
    \includegraphics[width=\linewidth]{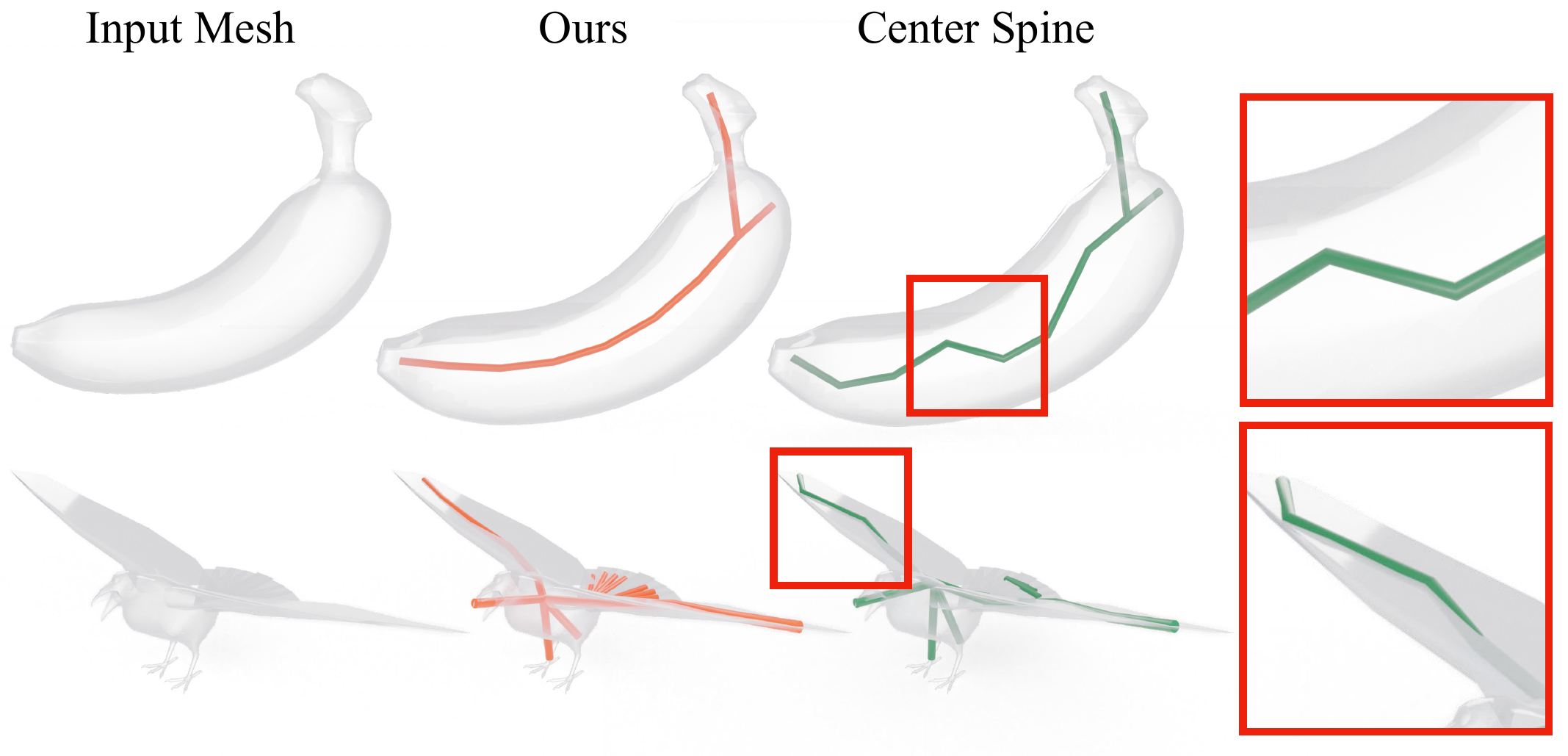}
    \caption{
    \textbf{Score-maximization spine vs.\ midplane center.}
    The midplane-center baseline often drifts away from the natural interior of the shape and produces a kinked spine trajectory, whereas our method maintains a smooth and geometrically consistent spine within the object.}
    \label{fig:ablation_spine}
    \vspace{-15pt}
\end{figure}

\paragraph{(b) Spine point extraction strategy.}
We compare our score-based spine point extraction (\S\ref{sec:spine_gen}) with a center baseline that computes the centroid of rib intersections with a midplane orthogonal to the principal axis. While both methods behave similarly on convex and axis-aligned parts, the center baseline often drifts outside the rib interior on non-convex or tilted geometry, producing visibly kinked and off-axis spine trajectories (Figure~\ref{fig:ablation_spine}). In contrast, our score-maximization strategy generates smoother and more stable spine structures, leading to more coherent spine-driven deformations and twist propagation.

\vspace{-10pt}

\begin{figure}[!t]
    \centering
    \includegraphics[width=\linewidth]{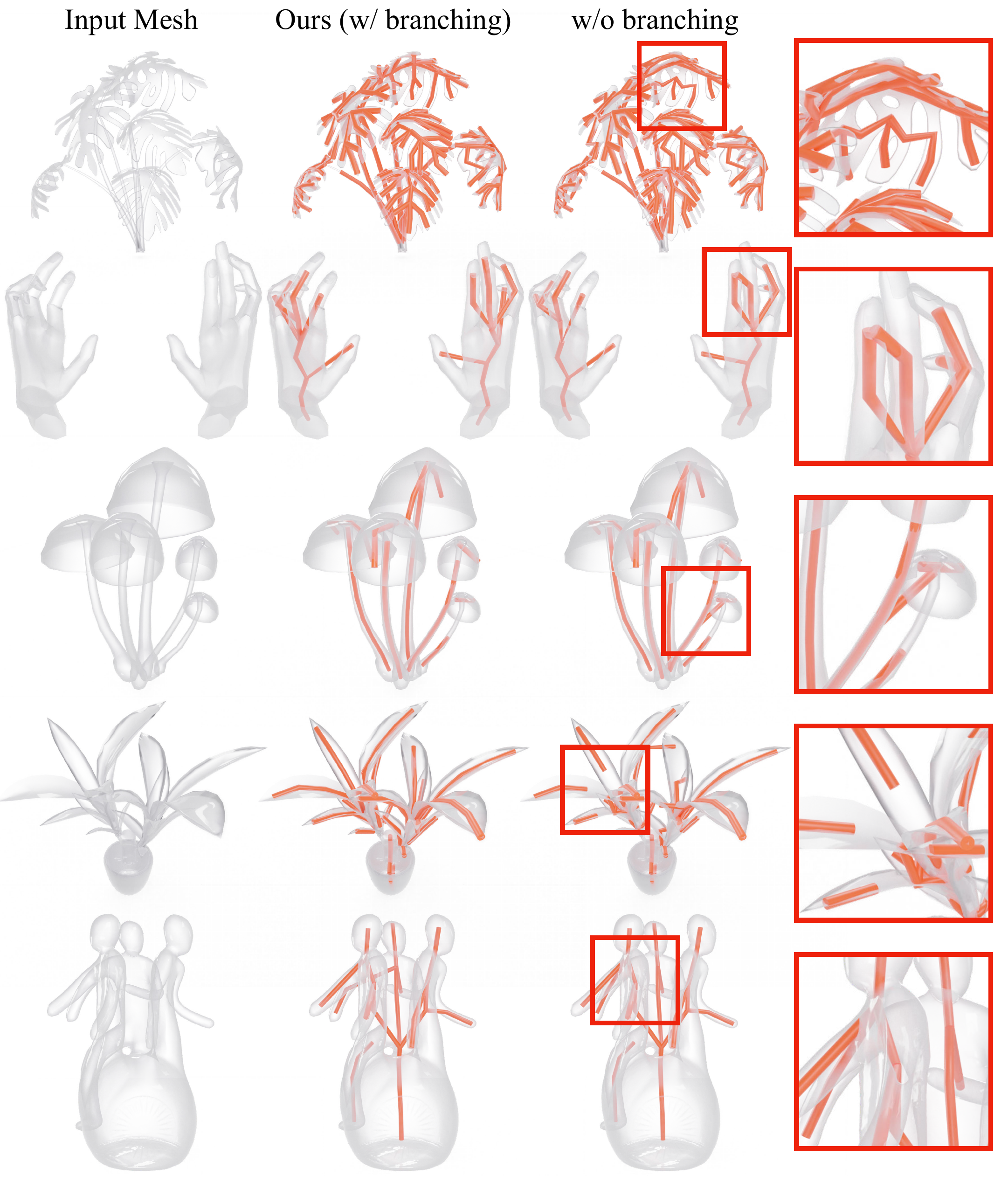}
    \caption{%
        \textbf{Branching tree handling.}
        Effect of the BFS-based parent-child sub-rib tracking on multi-limb shapes. The proposed topology-aware branching tree connects sub-ribs across consecutive iso-levels through the face-adjacency graph, enabling each branch spine to consistently follow its corresponding limb while sharing a single key point at Y-junctions. Without branching-tree handling, parent-child connectivity is removed and sub-ribs are instead matched across iso-levels using 3D nearest-neighbor search, causing the spine to jump between unrelated branches and collapse the per-limb structural consistency.
    }
    \label{fig:ablation_branch}
    \vspace{-15pt}
\end{figure}

\paragraph{(c) Branching tree handling.}
Branching tree handling provides explicit parent-child relationships between spine points, yielding a clear hierarchical structure and preventing incorrect merging across disconnected branches. Without branching tree handling, nearest-neighbor association across iso-levels fails on branched or multi-limb shapes. As shown in Figure~\ref{fig:ablation_branch}, spine points jump across disconnected sub-ribs, producing incoherent spine topology and unstable deformations.

\begin{figure}[!t]
    \centering
    \includegraphics[width=\linewidth]{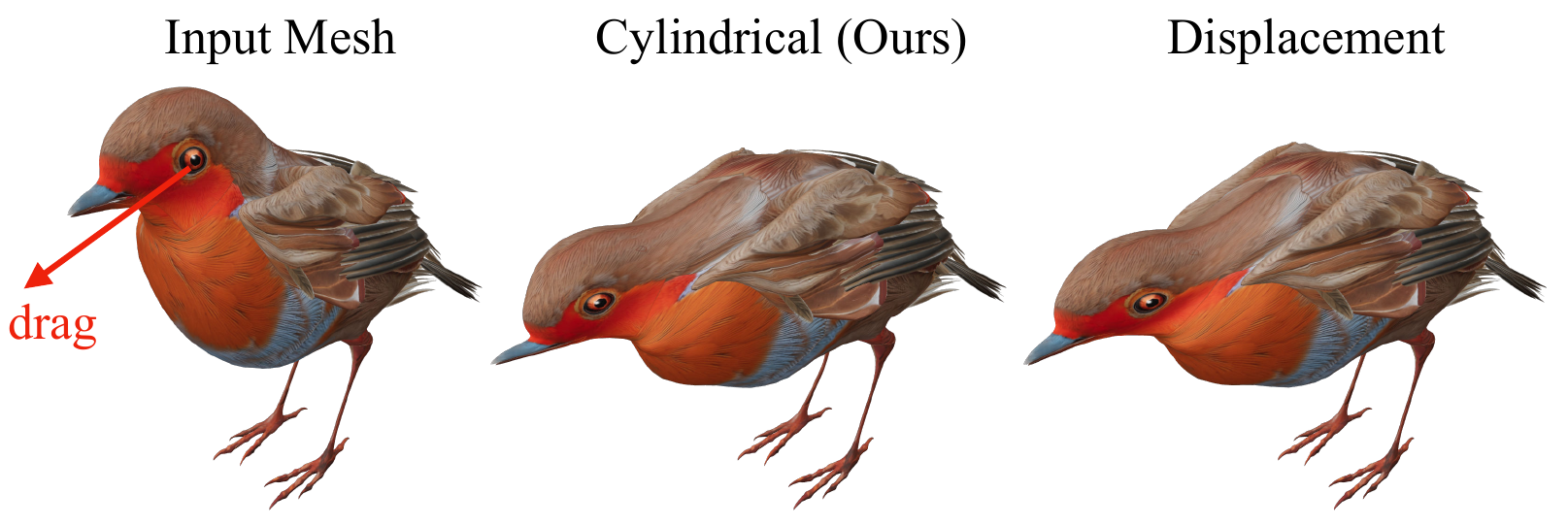}
    \caption{%
        \textbf{Mesh lifting.}
        Effect of the two mesh-lift modes on a drag-induced bending deformation that pulls the bird's
        head forward and downward.
        \emph{Input mesh:} the rest pose with the user drag (red arrow)
        applied at the head.
        \emph{Cylindrical (Ours):} per-vertex positions are reconstructed
        in the parallel-transport frame of the deformed spine segment,
        so rib cross-sections rotate together with the bending spine and
        the head retains its rounded volumetric profile.
        \emph{Displacement:} per-vertex displacement is a
        $\mathbf{W}^s$-weighted linear blend of sparse spine translations,
        so cross-sections stay parallel to their rest orientation while
        the spine bends, flattening the head.
    }
    \label{fig:ablation_meshlifting}
    \vspace{-10pt}
\end{figure}

\paragraph{(d) Mesh lifting.}
In reduced simulation, mesh lifting via cylindrical coordinates preserves rotational deformation throughout time integration. We ablate this design by replacing it with a per-vertex displacement lift. The per-vertex displacement lift blends sparse spine transformations through $\mathbf{W}_s$, while the cylindrical lift evaluates vertices in their rest cylindrical frame to better preserve local cross-sectional orientation and rotation during large motions. Across the $188$-mesh benchmark, the two lifting schemes agree closely on average, with a cross-mesh mean deviation of $0.46 \pm 0.48\%$, median $0.31\%$, and maximum $3.25\%$ of the bounding-box diagonal. In contrast, the per-mesh worst-vertex deviation has a cross-mesh mean of $9.43 \pm 10.06\%$, median $5.76\%$, and tail reaching $68.47\%$, mainly concentrated in strongly bending or twisting regions where the linear displacement lift cannot fully capture the cross-sectional rotations recovered by the cylindrical lift. As shown in Fig.~\ref{fig:ablation_meshlifting}, the cylindrical lift captures more plausible dynamics while better preserving the original head shape of the bird.

\section{Comparison and Application}
\label{sec:comparison_application}
\begin{figure*}[!t]
    \centering
    \includegraphics[width=\textwidth]{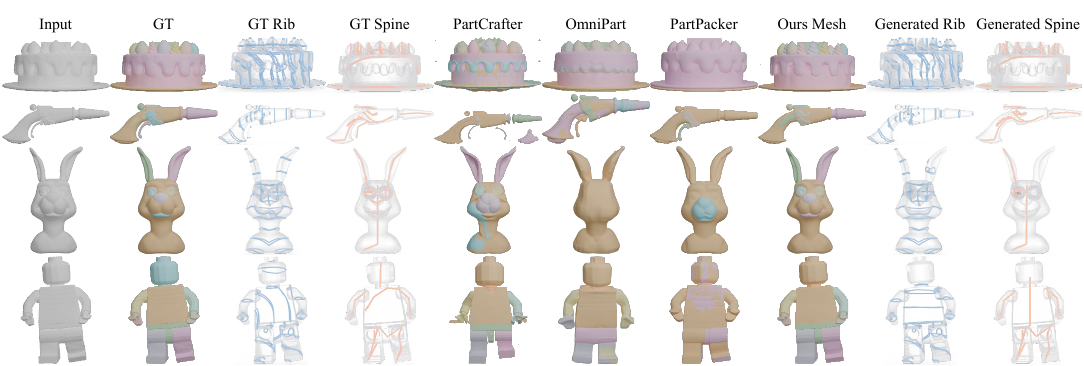}
    \caption{\textbf{Image-conditioned mesh + Fishbone generation.} The proposed method predicts the part-decomposed mesh and rib-spine structures from a single input image in a single forward pass. Compared with the baselines, the generated meshes exhibit cleaner part decomposition, more consistent geometric structures, and fewer floating artifacts.}
    \label{fig:fishbone_gen_demo}
\end{figure*}

We present four downstream applications of the proposed framework: image-conditioned joint generation of 3D shapes and rib-spine control structures, interactive mesh editing, robot-learning scene diversification, and closed-loop agentic deformation. We further evaluate the proposed method on these applications through comprehensive qualitative and quantitative comparisons. 

\subsection{3D Shape Generation}
\label{sec:app_gen}
\paragraph{Image-conditioned 3D and Fishbone generation.}

We extend PartCrafter \cite{lin2025partcrafter} into a unified feed-forward generator that jointly predicts a part-decomposed mesh together with the proposed rib-spine control structure from a single image. Each per-part latent consists of $K{=}1024$ tokens produced by PartCrafter's pretrained VAE, whose decoder predicts a truncated signed distance function (TSDF). The pretrained TSDF decoder remains frozen during training, while an additional autoregressive head is optimized to generate rib and spine tokens. Following UniRig~\cite{zhang2025one}, we employ an OPT-350M transformer as the autoregressive head. The per-part latent is projected through a linear layer and prepended to the token embeddings as a conditioning prefix, enabling autoregressive generation over a joint sequence containing both rib polylines and the spine structure. Since the rib-spine representation is conditioned on the same latent as the mesh generation branch, the predicted control structure remains geometrically consistent with the generated shape. We explore the following training loss for the autoregressive head:
\begin{equation}
    \mathcal{L} = \lambda_r \mathcal{L}_{\text{rib}} + \lambda_s \mathcal{L}_{\text{spine}},
\end{equation}
where both rib and spine terms are next-token-prediction (NTP) cross-entropies~\cite{zhang2025one}:
\begin{equation}
    \mathcal{L}_{\{\text{rib},\text{spine}\}} = -\sum_t \log P\bigl(\omega_t \mid \omega_{<t}, c\bigr),
\end{equation}
with $\omega_t$ denoting the autoregressive token at step $t$ and $c$ the per-part latent conditioning prefix. At inference time, starting from an image, the model produces a $(\mathcal{M},\mathcal{F})$ pair, enabling subsequent skinning-weight generation, controllable deformation, reduced-space dynamics, and animation.

We compare the proposed joint 3D mesh and Fishbone generation framework against representative baselines, including PartCrafter~\cite{lin2025partcrafter}, OmniPart~\cite{yang2025omnipart}, and PartPacker~\cite{tang2026efficient}. Folloing PartCrafter \cite{lin2025partcrafter}, We report Chamfer Distance and F1-score between the generated shapes and the ground-truth geometry as quantitative metrics. As shown in Tab.~\ref{tab:fishbone_gen_metrics}, our method achieves higher reconstruction accuracy than the baselines, as the predicted meshes are cleaner and contain fewer floating artifacts. While PartCrafter may produce incorrect part segmentations, OmniPart and PartPacker often generate under-segmented part decompositions. As shown in Fig.~\ref{fig:fishbone_gen_demo}, the jointly generated meshes and rib-spine structures exhibit cleaner geometry and fewer floating artifacts than the baselines, demonstrating that the rib-spine control representation provides geometrically consistent structural guidance during generation.

\begin{table}[t]
    \centering
    \footnotesize
    \setlength{\tabcolsep}{4pt}
    \setlength{\aboverulesep}{0pt}
    \setlength{\belowrulesep}{0pt}
    \begin{tabular*}{\linewidth}{@{\extracolsep{\fill}}l|cccc@{}}
        \toprule
        \textbf{Metric}
            & PartCrafter~\cite{lin2025partcrafter}
            & OmniPart~\cite{yang2025omnipart}
            & PartPacker~\cite{tang2026efficient}
            & \textbf{Ours} \\
        \midrule
        \textbf{CD}~$\downarrow$
            & $0.3085$ & $0.2786$ & $0.2783$ & \cellcolor{bestcolor}$0.0723$ \\
        \textbf{F-Score}~$\uparrow$
            & $0.5861$ & $0.5884$ & $0.6048$ & \cellcolor{bestcolor}$0.9118$ \\
        \bottomrule
    \end{tabular*}
    \caption{
        \textbf{Image-conditioned generation comparison.}
        Quantitative results using Chamfer Distance (CD) and F-Score.
    }
    \label{tab:fishbone_gen_metrics}
    \vspace{-10pt}
\end{table}

\subsection{3D Shape Editing}
\label{sec:app_edit}
\begin{table}[t]
    \centering
    \footnotesize
    \setlength{\tabcolsep}{2pt}
    \setlength{\aboverulesep}{0pt}
    \setlength{\belowrulesep}{0pt}
    \begin{tabular*}{\linewidth}{@{\extracolsep{\fill}}l|l|ccc|ccc@{}}
        \toprule
        \textbf{Mesh} & \textbf{Method}
            & $\mathrm{Disp}_{m}$
            & $\mathrm{Edge}_{m}$
            & $\mathrm{Area}_{m}$
            & $\mathrm{Lap}_{\max}$~$\downarrow$
            & ARAP $E$~$\downarrow$
            & Prep.~$\downarrow$ \\
        \midrule
        \multirow{3}{*}{Dragon}
            & ARAP          & $11.38$ & $43.88$ & $68.82$ & $3.68$ & $4.45$ & N/A \\
            & WIR3D         & $0.91$  & $19.48$ & $28.53$ & $65.81$ & $24.40$ & $4.46$~h \\
            & \textbf{Ours} & $11.54$ & $42.48$ & $66.93$ & \cellcolor{bestcolor}$2.69$ & \cellcolor{bestcolor}$4.34$ & \cellcolor{bestcolor}$17.95$~s \\
        \midrule
        \multirow{3}{*}{Nefertiti}
            & ARAP          & $12.41$ & $44.65$ & $69.31$ & $4.72$ & $3.86$ & N/A \\
            & WIR3D         & $0.98$  & $14.72$ & $24.51$ & $108.18$ & $15.72$ & $9.86$~h \\
            & \textbf{Ours} & $12.21$ & $44.44$ & $69.12$ & \cellcolor{bestcolor}$1.94$ & \cellcolor{bestcolor}$3.78$ & \cellcolor{bestcolor}$10.02$~s \\
        \midrule
        \multirow{3}{*}{Chair}
            & ARAP          & $23.24$ & $110.20$ & $496.39$ & $2703.90$ & $10.65$ & N/A \\
            & WIR3D         & $13.46$ & $154.41$ & $143.08$ & $1683.38$ & $10.09$ & $3.73$~h \\
            & \textbf{Ours} & $23.29$ & $89.05$ & $256.43$ & \cellcolor{bestcolor}$93.37$ & \cellcolor{bestcolor}$1.14$ & \cellcolor{bestcolor}$12.00$~s \\
        \midrule
        \multirow{3}{*}{Spot}
            & ARAP          & $7.07$ & $78.35$ & $100.36$ & $30.38$ & $25.14$ & N/A \\
            & WIR3D         & $8.95$  & $86.69$ & $112.78$ & $63.54$ & $53.57$ & $8.59$~h \\
            & \textbf{Ours} & $6.89$  & $73.70$ & $94.92$ & \cellcolor{bestcolor}$23.61$ & \cellcolor{bestcolor}$20.50$ & \cellcolor{bestcolor}$2.21$~s \\
        \bottomrule
    \end{tabular*}
    \caption{%
        \textbf{Static deformation comparison.}
        Per-mesh deformation-quality metrics for ARAP~\cite{sorkine2007rigid}, WIR3D~\cite{Liu_2025_ICCV}, and \textbf{Ours} on a small set of in-the-wild meshes, all driven toward a comparable target deformation.
        $\mathrm{Disp}_{m}$ (mean per-vertex displacement as a percentage of the bounding-box diagonal), $\mathrm{Edge}_{m}$ (mean relative edge-length change), and $\mathrm{Area}_{m}$ (mean relative triangle-area change between rest and deformed mesh) are descriptive measures of the deformation each method produced, characterizing its magnitude and structural distortion, with no inherent better direction.
        $\mathrm{Lap}_{\max}$ (lower $=$ fewer isolated artifacts) is the worst-case Laplacian-smoothness deviation in percent, ARAP $E$ (lower $=$ closer to local rigidity) is the as-rigid-as-possible energy of the deformed mesh, and Prep.\ is the preprocessing time (plus any manual handle authoring) needed before the first edit can be issued, reported in seconds (s) for sub-minute timings and hours (h) for the rest.
        The pink cell highlight marks where \textbf{Ours} is the best among the three methods on the quality columns ($\mathrm{Lap}_{\max}$ and ARAP $E$) and the lowest measured preprocessing time. The first three columns are not highlighted since they do not have an inherent better direction.
    }
    \label{tab:comparison}
    \vspace{-10pt}
\end{table}
\begin{figure*}[t]
    \centering
    \includegraphics[width=\textwidth]{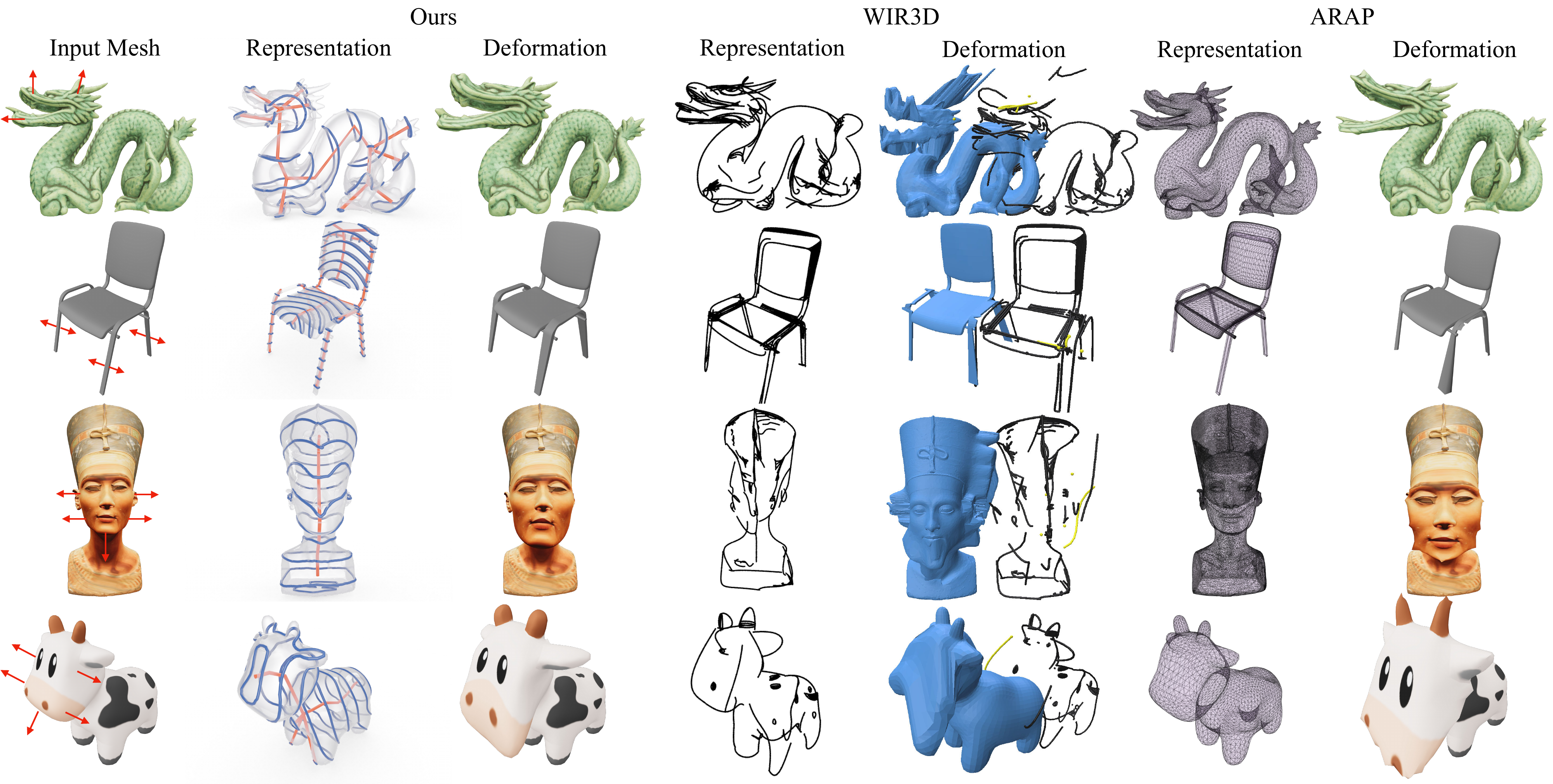}
    \vspace{-20px}
    \caption{%
        \textbf{Qualitative editing comparison.}
        Side-by-side editing results of WIR3D~\cite{Liu_2025_ICCV}, ARAP~\cite{sorkine2007rigid}, and \textbf{Fishbone}. \textbf{Fishbone} provides explicit rib-spine handles that support localized and structured edits with lower manual setup.
    }
    \label{fig:comparison_editing}
\end{figure*}

\begin{table}[!h]
    \centering
    \footnotesize
    \setlength{\tabcolsep}{3pt}
    \begin{tabular*}{\linewidth}{@{\extracolsep{\fill}}lcccccc@{}}
        \toprule
        & \multicolumn{2}{c}{\textbf{Zero-shot}} & \multicolumn{2}{c}{\textbf{+\,Fishbone}} & \multicolumn{2}{c}{$\boldsymbol{\Delta}$} \\
        \cmidrule(lr){2-3} \cmidrule(lr){4-5} \cmidrule(lr){6-7}
        \textbf{Category} & Single & Multi & Single & Multi & Single & Multi \\
        \midrule
        Croissant & $37.27$ & $11.88$ & $48.05$ & $26.89$ & $+10.77$ & $+15.01$ \\
        Tuna      & $21.72$ & $6.800$ & $26.89$ & $24.17$ & $+5.170$ & $+17.38$ \\
        Cup       & $87.42$ & $61.17$ & $89.49$ & $72.76$ & $+2.060$ & $+11.59$ \\
        \bottomrule
    \end{tabular*}
    \caption{%
        \textbf{Dexterous grasping on in-the-wild assets.}
        Lift-and-hold success rate (\%) of DexGraspNet~2.0~\cite{zhang2024dexgraspnet} on held-out in-the-wild meshes, in single-object and multi-object scenes.
        \emph{Zero-shot} uses the released checkpoint as-is. \emph{+\,Fishbone} fine-tunes it on \textbf{Fishbone}-generated deformation variants seeded from a small in-category mesh library. $\Delta$ reports the absolute percentage-point improvement over the zero-shot baseline.
    }
    \label{tab:robot_results}
    \vspace{-25pt}
\end{table}
\textbf{Fishbone} provides explicit rib-spine parameters for intuitive mesh editing: users can thicken a plant stem by scaling ribs, bend a lamp neck through spine rotation, or twist a chair backrest. We further provide an interactive GUI where users select rib or spine handles and adjust deformation parameters with sliders. Edits are applied at interactive rates using pre-cached skinning weights (Fig.~\ref{fig:gui} in the appendix).We compare \textbf{Fishbone}'s mesh deformation against two representative baselines: as-rigid-as-possible (ARAP) deformation~\cite{sorkine2007rigid}, a canonical energy-minimization framework that propagates user constraints through a local rigidity prior, and WIR3D~\cite{Liu_2025_ICCV}, a recent geometry-aware shape abstraction that represents meshes using a sparse set of 3D wires serving as deformation handles. We evaluate on four in-the-wild meshes (Dragon, Nefertiti, Chair, and Spot) spanning organic, free-form, and man-made shapes. For each mesh and method, we author a comparable target deformation and explore six evaluation metrics on the deformed results. Three metrics assess deformation magnitude, including \emph{(i)} $\mathrm{Disp}_{m}$ (\%), the mean per-vertex displacement normalized by the bounding-box diagonal, \emph{(ii)} $\mathrm{Edge}_{m}$ (\%), and \emph{(iii)} $\mathrm{Area}_{m}$ (\%), the mean relative edge-length and triangle-area changes. The remaining three assess deformation quality and structural consistency, including \emph{(iv)} $\mathrm{Lap}_{\max}$ (\%) $\downarrow$, the maximum Laplacian smoothness deviation capturing local artifacts, \emph{(v)} ARAP energy $\downarrow$~\cite{sorkine2007rigid}, measuring local rigidity preservation, and \emph{(vi)} preprocessing time $\downarrow$, including both runtime and manual preparation cost. Lower values indicate better deformation quality and lower preparation overhead. Our method and ARAP produce deformations of comparable magnitude, with similar $\mathrm{Disp}_{m}$, $\mathrm{Edge}_{m}$, and $\mathrm{Area}_{m}$ ranges, while \textbf{Fishbone} consistently achieves lower $\mathrm{Lap}_{\max}$ and ARAP energy (Tab.~\ref{tab:comparison}) at a fraction of the preprocessing cost (Prep.\ of $2$-$18$ seconds vs.\ WIR3D's $3.73$-$9.86$ hours). 
In addition, WIR3D shows inconsistent deformation quality across meshes: $\mathrm{Disp}_{m}$ remains below $1\%$ on Dragon and Nefertiti (vs.\ $\sim12\%$ for ARAP / \textbf{Fishbone}), while Chair and Spot achieve comparable deformation magnitudes but with severe local artifacts (\mbox{$\mathrm{Lap}_{\max}$} up to $1683\%$ and ARAP energy up to $\sim5\times$ higher than \textbf{Fishbone}). ARAP preserves local rigidity once handles are authored, but often struggles with free-form non-rigid edits, whereas WIR3D relies on category-specific abstraction training and careful wire selection. As shown in Fig.~\ref{fig:comparison_editing}, ARAP tends to amplify distortion when the rigidity prior conflicts with the intended deformation, while WIR3D introduces inconsistent local artifacts. In comparison, ours achieves smoother surface behavior and fewer local artifacts under comparable target deformations, offering smoother and more stable deformation behavior under comparable target edits.

\begin{figure}[!h]
    \centering
    \vspace{-10px}\includegraphics[width=\linewidth]{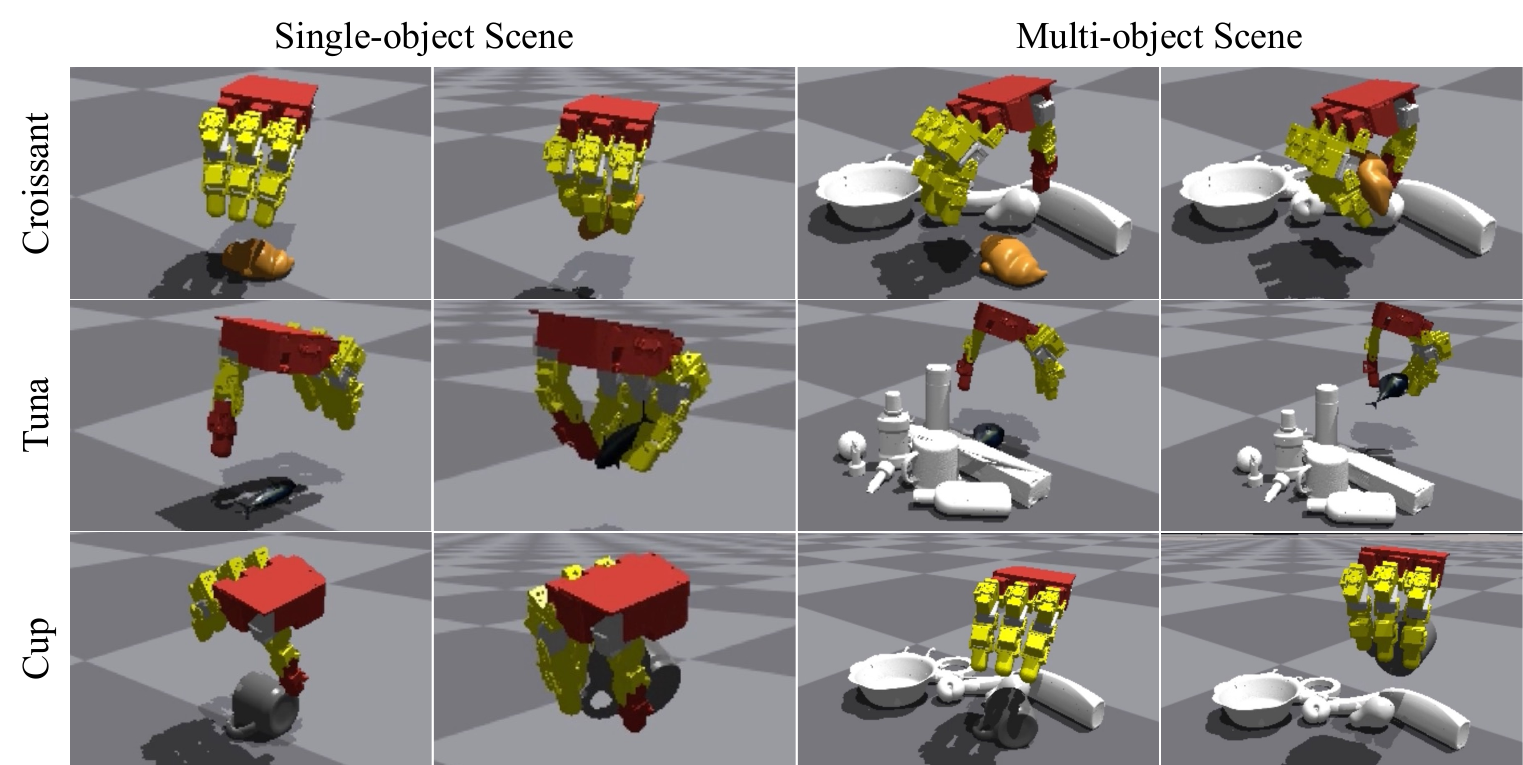}
    \caption{%
        \textbf{Dexterous grasping on in-the-wild scenes.}
        Representative DexGraspNet~2.0~\cite{zhang2024dexgraspnet} grasp executions on unseen in-the-wild meshes from three categories (rows, top to bottom: Croissant, Tuna, Cup). The left two columns show single-object grasp scenes before and after execution, while the right two columns show the corresponding multi-object grasp scenes with category-agnostic clutter surrounding the target. The proposed data diversification enables stable and accurate grasps under both single-object and cluttered multi-object settings.
    }
    \label{fig:dex_grasp_demos}
    \vspace{-20pt}
\end{figure}

\begin{figure*}[!h]
    \centering
    \includegraphics[width=\textwidth]{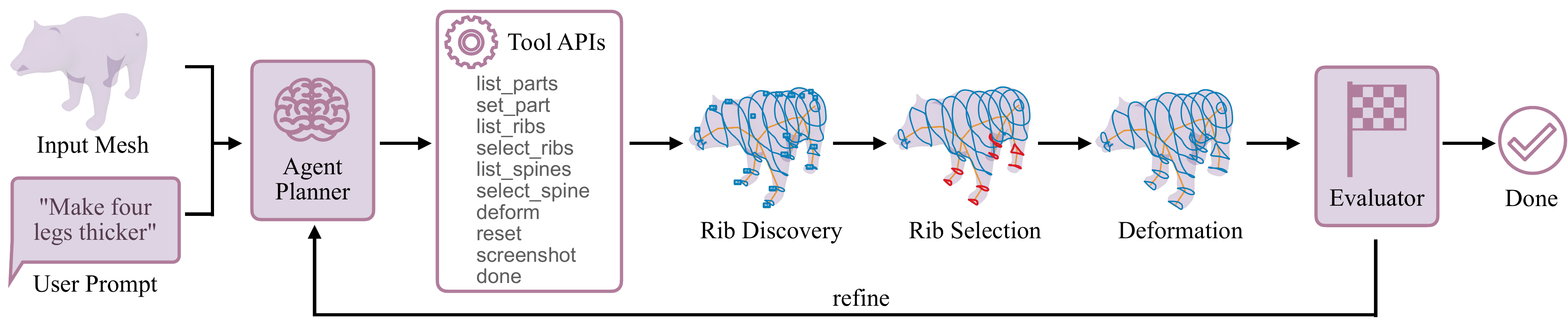}
    \vspace{-10pt}
    \caption{\textbf{Closed-loop Fishbone agent.} The VLM observes a rendered screenshot, picks a tool from the tools in Tab.~\ref{tab:agent_tools}, applies it, and re-renders for the next round. An evaluator critic judges every \texttt{deform} call against the original request to gate retries.}
    \label{fig:agent_pipeline}
    \vspace{-5pt}
\end{figure*}

\subsection{Robot Learning via Data Diversification}
\label{sec:app_robot}

We use \textbf{Fishbone} as a geometry-aware data diversification framework for dexterous grasping. Specifically, we adopt DexGraspNet~2.0~\cite{zhang2024dexgraspnet} as the grasping policy, which predicts a Shadow-Hand grasp pose from a single-view scene point cloud. We compare the original training data with \textbf{Fishbone}-augmented data by applying the proposed rib- and spine-driven deformations to produce 500 geometrically diverse object variants per object while preserving the semantic structure of the original assets. By sampling deformations such as per-rib scaling, cross-section reshape, spine bending, and twisting, \textbf{Fishbone} produces a broader range of object thicknesses, curvatures, aspect ratios, and local shape variations for within-category data augmentation. We then fine-tune DexGraspNet~2.0 on scenes constructed from this augmented pool without changing the network architecture. Since the test meshes are disjoint from both the seed set and the deformation process, the evaluation measures out-of-distribution generalization rather than memorization. As shown in Tab.~\ref{tab:robot_results}, \textbf{Fishbone} consistently improves grasp success in both single-object and multi-object scenes, highlighting the importance of object-level geometric diversity for manipulation generalization~\cite{tobin2017domain, peng2018sim}. Qualitative results in Fig.~\ref{fig:dex_grasp_demos} further show that the augmented training data enables more robust grasping on challenging in-the-wild objects and cluttered scenes.

\subsection{Agentic Mesh Deformation}
\label{sec:app_agent}

\begin{figure}[h]
    \centering
    \includegraphics[width=\linewidth]{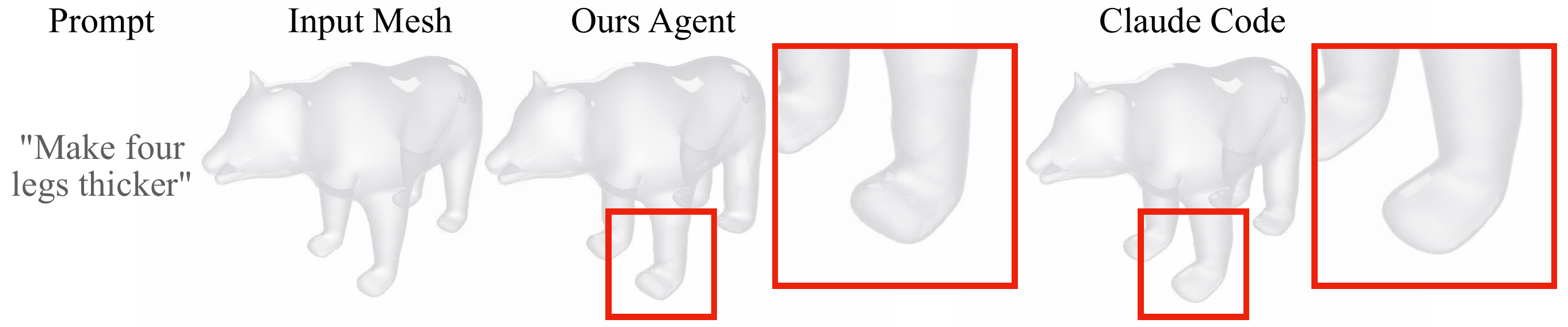}
    \caption{\textbf{Agentic editing comparison.} Editing results of \textbf{Fishbone} and Claude Code from prompts. The proposed rib-spine handles keep each edit on-target and geometrically plausible, while Claude Code agents uniformly scale large portions of the mesh instead of the requested local editing.}
    \label{fig:comparison_agentic}
    \vspace{-10pt}
\end{figure}

Given the explicit rib-spine control representation extracted by the proposed method, we further explore agentic mesh deformation through VLM agents. We build a closed-loop \textbf{Fishbone} agent that converts natural-language requests into mesh deformations (Fig.~\ref{fig:agent_pipeline}). Each rib and spine branch carries a stable geometry-derived semantic name, while every deformation primitive is represented by low-dimensional and semantically meaningful control parameters, enabling an LLM to reason about edits such as ``thickening the four legs'' or ``bending the body forward'' without per-shape fine-tuning. The VLM-driven agent continuously observes rendered screenshots and predicts the next tool call from a ten-tool API (Tab.~\ref{tab:agent_tools}), including part / rib / spine-branch selection, deformation primitives, \texttt{reset}, \texttt{screenshot}, and \texttt{done}. Each call returns text together with a new rendering, forming a visual feedback loop for iterative refinement. We further introduce a VLM evaluator that verifies whether each deformation satisfies the user request by measuring the semantic consistency between the input prompt and the rendered deformation result. We further demonstrate a prototype closed-loop VLM agent built on Fishbone controls, where the resulting edits remain semantically coherent and geometrically plausible as indicated in Fig.~\ref{fig:comparison_agentic}. 

\section{Conclusion and Future Work}
\label{sec:conclusion}
We presented \textbf{Fishbone}, a novel framework for representing and deforming 3D shapes through a coupled rib-spine control structure automatically extracted from general meshes via the proposed adaptive heat method and robust spine-centering procedure, without requiring manual annotation, template fitting, or per-shape rigging. The resulting representation supports six rib-driven and three spine-driven deformation primitives, enabling accurate real-time local and global shape deformation within a unified parameter space. Moreover, the same representation is simulation-ready and naturally supports reduced-space real-time dynamics and keyframe animation. The proposed method further supports diverse downstream applications, including agentic mesh editing, controllable asset processing, image-conditioned 3D generation, and robot learning. 

\paragraph{Limitations and future work.}
While \textbf{Fishbone} provides an automatic and geometry-aware handle system for controllable deformation and reduced-space dynamics, several limitations remain for future work. First, the current rib-spine representation is defined only on triangle meshes. It is possible to integrate the proposed control structures with more 3D representations. For example, point-cloud variants could replace iso-geodesic contours with cross-sectional distance bands to inherit similar controllable handles without explicit meshing, while integration with 3D Gaussian Splatting and NeRF may enable controllable deformation and animation directly within radiance-field representations. Second, keyframe authoring still requires users to manually compose animations through \textbf{Fishbone} edits. Although intuitive, this becomes labor intensive for long sequences. A learned keyframe generator conditioned on text prompts, sparse trajectories, or exemplar videos could directly predict motion parameters and simplify animation creation.

\bibliographystyle{ACM-Reference-Format}
\bibliography{references}

\clearpage
\appendix
\section{Mesh Preprocessing}
\label{sec:preprocess}

This appendix details the input-cleaning and watertightness-repair stages that precede the rib/spine extraction of \S\ref{sec:rib_gen}.

\paragraph{Cleaning.}
Meshes from 3D generation pipelines, scanning, or artist tools are often not directly usable: they contain NaN vertices, zero-area faces, near-duplicate vertices, disjoint parts, and arbitrary scale and orientation.
We apply a lightweight cleaning pass that (i) removes NaN and degenerate faces, (ii) merges vertices within tolerance $\epsilon_{\text{merge}} = 10^{-6}$, (iii) optionally keeps only the largest connected component, and (iv) normalizes the mesh to a unit bounding box centered at the origin.
For multi-part objects (e.g., GLB scenes with multiple geometry nodes), each part is extracted and passed through the pipeline independently, while a shared normalization transform preserves the inter-part spatial relationships.

\paragraph{Watertightness repair.}
Many AI-generated or scanned parts that are intended to be solid remain non-watertight after cleaning, with sub-visible cracks that would spuriously split otherwise continuous iso-contours into broken open polylines.
When a part fails the watertightness test \emph{and} is classified as solid by the sheet detector of \S\ref{sec:rib_gen}, we generate a \emph{welded geometric twin}: a second copy of the part in which spatially coincident vertices are aggressively merged (tolerance $\propto$ bbox diagonal $\cdot\,10^{-4}$), small boundary holes are closed by hole filling, and duplicate / degenerate faces are removed.
Genuine thin-shell meshes (sheets, fins, leaves) are detected separately (\S\ref{sec:rib_gen}) and bypass this repair, so their authentic boundary edges and the open ribs they induce are preserved.
The original part, with its original UV layout and textures, is kept unchanged for downstream rendering, while the welded twin is consumed by the geodesic stage.
After skinning weights are computed on the twin, we expand every per-vertex and per-nonzero array back to the original vertex count using a KD-tree inverse map, so that the resulting $\mathbf{W}$ is consistent with the original mesh connectivity.

\section{GPU Acceleration and Caching}
\label{sec:gpu_accel}

This appendix details the implementation of the per-rib nearest-point computation referenced at the end of \S\ref{sec:weights_gen}, which dominates runtime on dense meshes.

\paragraph{KD-tree on CPU.}
We build a KD-tree over the union of all rib polyline vertices and query it with a per-vertex radius set to the distance at which the raw Gaussian falls to $w_{\min}$ (i.e., $\sigma\sqrt{-2\ln w_{\min}}$ for the adaptive bandwidth $\sigma$ of \S\ref{sec:weights_gen}). Inside that ball we vectorize point-to-segment distances over all candidate rib edges and pick the minimum-distance edge per vertex.
This avoids the quadratic cost of brute-force enumeration while remaining exact within the support radius.

\paragraph{Batched tensor kernel on GPU.}
On GPU we replicate the same math as a batched tensor kernel: for each part, the per-vertex point-to-segment distance is evaluated against all rib edges in parallel using a $(N \times K \times M)$ broadcast (vertices, ribs, segments per rib), with masking for invalid edges and for vertices outside the soft-cutoff radius.
The kernel produces both the nearest-edge index and the parametric position $t_{ik}$ in a single pass, feeding directly into Eq.~\eqref{eq:soft_weight}.
For multi-part assets we batch parts in parallel via process pools that pin one part to one CUDA device.

\paragraph{Disk caching.}
Recomputing $\mathbf{W}$ for an already-processed asset is wasted work: weights depend deterministically on the cleaned mesh, the rib polylines, and the bandwidth schedule.
We therefore cache $\mathbf{W}$, the per-vertex projection metadata, and the per-spine-key-point spine-weight matrices on disk, keyed by a SHA-256 hash of the cleaned vertex/face buffers concatenated with the rib/spine bytes. Subsequent deformation sessions load the cache in milliseconds and bypass the weight stage entirely.
Cache invalidation is automatic on any upstream change because the hash inputs cover every quantity the weights depend on.

\paragraph{Batch throughput.}
Dispatching one part per worker through a process pool with one worker per GPU, processing the $2{,}565$-mesh extraction run reported in \S\ref{sec:performance} accumulates $216{,}687.7$~s ($60.2$~h) of single-worker runtime, divided in practice across the workstation's eight H100 NVL GPUs, and yields the \textbf{Fishbone}-augmented dataset introduced as a contribution in \S1.

\section{Spine-Dynamics Force Gradients}
\label{sec:dyn_force_gradients}

This appendix lists the per-key-point force terms used by the integrator of Eq.~\ref{eq:dyn_integrator}.
Each term is the negative gradient of the corresponding energy in Eqs.~\ref{eq:dyn_stretch}-\ref{eq:dyn_bend}, accumulated into a force buffer $\mathbf{f} \in \mathbb{R}^{K \times 3}$ before the velocity update.

\paragraph{Rest-length-scaled stretch.}
Let $\mathbf{e}_i = \mathbf{p}_{i+1} - \mathbf{p}_i$, $\ell_i = \|\mathbf{e}_i\|$, $\hat{\mathbf{e}}_i = \mathbf{e}_i / (\ell_i + \varepsilon)$, and $\sigma_i = \bar\ell^0 / (\ell_i^0 + \varepsilon)$ for every within-branch edge.
The axial-spring force from $-\nabla E_{\text{stretch}}$ (Eq.~\ref{eq:dyn_stretch}) is
\begin{subequations}
\label{eq:dyn_force_stretch}
\begin{align}
    \mathbf{f}_i^{\text{stretch}}     &\mathrel{+}= k_s \, \sigma_i \, (\ell_i - \ell_i^0) \, \hat{\mathbf{e}}_i, \\
    \mathbf{f}_{i+1}^{\text{stretch}} &\mathrel{+}= -k_s \, \sigma_i \, (\ell_i - \ell_i^0) \, \hat{\mathbf{e}}_i.
\end{align}
\end{subequations}

\paragraph{Length-normalized curvature bending.}
For every within-branch triple $(i-1, i, i+1)$, let $\mathbf{e}_{i-1} = \mathbf{p}_i - \mathbf{p}_{i-1}$, $\mathbf{e}_i = \mathbf{p}_{i+1} - \mathbf{p}_i$, with current lengths $\ell_{i-1} = \|\mathbf{e}_{i-1}\|$, $\ell_i = \|\mathbf{e}_i\|$ and tangents $\mathbf{t}_{i-1} = \mathbf{e}_{i-1} / \ell_{i-1}$, $\mathbf{t}_i = \mathbf{e}_i / \ell_i$.
Form the discrete curvature $\boldsymbol{\kappa}_i = 2(\mathbf{t}_i - \mathbf{t}_{i-1}) / L^0_i$ with $L^0_i = \ell^0_{i-1} + \ell^0_i$, the curvature residual $\Delta\boldsymbol{\kappa}_i = \boldsymbol{\kappa}_i - \boldsymbol{\kappa}_i^0$, and the per-triple coefficient $A_i = 2 k_b \bar\ell_i^0 / L^0_i$.
Using the projector $\mathbf{P}_e = \mathbf{I} - \mathbf{t}_e \mathbf{t}_e^\top$ that comes from $\partial(\mathbf{e}/\|\mathbf{e}\|)/\partial \mathbf{e} = \mathbf{P}_e / \|\mathbf{e}\|$, the force from $-\nabla E_{\text{bend}}$ (Eq.~\ref{eq:dyn_bend}) is
\begin{subequations}
\label{eq:dyn_force_bend}
\begin{align}
    \mathbf{f}_{i-1}^{\text{bend}} &\mathrel{+}= -\frac{A_i}{\ell_{i-1}} \, \mathbf{P}_{i-1} \, \Delta\boldsymbol{\kappa}_i, \\
    \mathbf{f}_{i+1}^{\text{bend}} &\mathrel{+}= -\frac{A_i}{\ell_i} \, \mathbf{P}_i \, \Delta\boldsymbol{\kappa}_i, \\
    \mathbf{f}_{i}^{\text{bend}}   &\mathrel{+}= -\bigl(\mathbf{f}_{i-1}^{\text{bend}} + \mathbf{f}_{i+1}^{\text{bend}}\bigr).
\end{align}
\end{subequations}
Newton's third law on the third line keeps total spine momentum exactly preserved per triple.
The codebase additionally retains the simpler rest-Laplacian bending force, $\mathbf{f}_{i\pm 1} \mathrel{+}= -k_b\,\mathbf{r}_i$, $\mathbf{f}_i \mathrel{+}= 2 k_b \mathbf{r}_i$ with $\mathbf{r}_i = \mathbf{p}_{i-1} - 2\mathbf{p}_i + \mathbf{p}_{i+1} - \Delta_i^0$, as a regression-test fallback.

\paragraph{Rayleigh damping.}
The mass-proportional and two stiffness-like damping kernels described in \S\ref{sec:dyn_state} accumulate as
\begin{subequations}
\label{eq:dyn_force_damp}
\begin{align}
    \mathbf{f}_i^{\text{damp},\alpha} &\mathrel{+}= -\alpha \, m_i \, \dot{\mathbf{p}}_i, \\
    \mathbf{f}_i^{\text{damp},e}      &\mathrel{+}= +\beta_e \, \bigl((\dot{\mathbf{p}}_{i+1} - \dot{\mathbf{p}}_i) \cdot \hat{\mathbf{e}}_i\bigr) \, \hat{\mathbf{e}}_i, \quad
    \mathbf{f}_{i+1}^{\text{damp},e}  \mathrel{+}= -\bigl(\text{same}\bigr), \\
    \mathbf{f}_{i\pm 1}^{\text{damp},b} &\mathrel{+}= -\beta_b \, \dot{\mathbf{d}}_i, \quad
    \mathbf{f}_{i}^{\text{damp},b}    \mathrel{+}= 2\beta_b \, \dot{\mathbf{d}}_i,
\end{align}
\end{subequations}
where $\dot{\mathbf{d}}_i = \dot{\mathbf{p}}_{i-1} - 2\dot{\mathbf{p}}_i + \dot{\mathbf{p}}_{i+1}$.
Setting $\beta_e = \beta_b = 0$ recovers a single mass-proportional kernel, and the legacy uniform form $\mathbf{f}_i^{\text{damp}} \mathrel{+}= -d\,\dot{\mathbf{p}}_i$ corresponds to the further reduction $m_i \equiv 1$.

\paragraph{Wind, gravity, impulse, projection.}
The drag-form wind of Eq.~\ref{eq:dyn_wind_full}, gravity (\S\ref{sec:dyn_forces}), the impulse update of Eq.~\ref{eq:dyn_impulse}, and the mesh-projected pull-back (\S\ref{sec:dyn_forces}) are added directly as written without further per-key-point bookkeeping.

\section{Default Parameters}
\label{sec:default_params}

This appendix consolidates the constants that the static and dynamic pipelines use throughout the paper.
They are all set once and left untouched across every result we report. Per-experiment overrides, when used, are flagged in the relevant subsection.

\paragraph{Mesh-to-rib-spine pipeline.}
Rib counts and influence bandwidths are auto-derived per part rather than fixed globally.
The number of ribs $K$ is scaled to each part's relative bounding-box extent and clipped to $[K_{\min}, K_{\max}] = [3, 10]$ (\S\ref{sec:rib_gen}).
The Gaussian skinning bandwidth is set to $\sigma = n\Delta/\sqrt{-2\ln w_{\min}}$ (\S\ref{sec:weights_gen}) with neighbor count $n = 2$, floor $w_{\min} = 10^{-4}$, and per-part rib spacing $\Delta = L_p/K$.
The 2D score grid for spine-point construction (\S\ref{sec:spine_gen}) uses $G = 128$ with scoring weights $(\alpha, \beta, \gamma) = (1.0, 0.0, 0.0)$ (flatness-only by default, with centroid-proximity and parent-proximity exposed as optional knobs).
The heat-method root axis is auto-selected from each part's AABB (\S\ref{sec:rib_gen}).
Meshes are normalized to a unit bounding box centered at the origin before processing. The watertightness-repair tolerance is $10^{-4}\,\cdot$\,bbox-diagonal (\S\ref{sec:preprocess}).

\paragraph{Reduced-coordinate dynamics.}
The integrator runs semi-implicit Euler with four substeps per frame (Eq.~\ref{eq:dyn_integrator}).
For the elastic potential we use $k_s$ and $k_b$ as user-facing dials, applied through the rest-length-scaled stretch (Eq.~\ref{eq:dyn_stretch}) and the length-normalized curvature bending (Eq.~\ref{eq:dyn_bend}). The per-part normalization $\sigma_i$ and $\bar\ell_i^0$ are computed from the rest spine.
The Rayleigh damping coefficients $(\alpha, \beta_e, \beta_b)$ in \S\ref{sec:dyn_state} default to $(\alpha, 0, 0)$, i.e.\ mass-proportional only. The edge and bending kernels are turned on per scene when oscillations need extra suppression.
For the wind models, the procedural ramp uses $p = 1.5$ in $a_i = (d_i / d_{\max})^p$ and the drag form (Eq.~\ref{eq:dyn_wind_full}) defaults to turbulence strength $\tau = 0.3$, secondary-frequency ratio $2.3$, and the tangent-perpendicular projector enabled.
The cylindrical mesh lift (Eq.~\ref{eq:dyn_cyl}) defaults its blend bandwidth to $\sigma_s = 0.05 \cdot \mathrm{bbox\text{-}diagonal}$, so vertices within roughly $5\%$ of the bbox follow the spine frame and farther vertices fall back to the displacement lift.
The user-impulse Gaussian uses the same per-mesh default unless overridden by the interaction layer.

\section{Editor GUI}
\label{sec:gui_app}

We provide a desktop editor for browsing and modifying the rib-spine handles of a \textbf{Fishbone}-rigged mesh (\S\ref{sec:app_edit}).
The editor displays the input mesh together with its ribs and spine. The user picks a rib or spine-branch handle from the side panel and adjusts its parameters with sliders, invoking the deformation primitives of \S\ref{sec:rib_deformation} in real-time via the pre-cached skinning matrix $\mathbf{W}$ (\S\ref{sec:weights_gen}), with no per-edit recomputation.

\begin{figure}[!h]
    \centering
    \includegraphics[width=\linewidth]{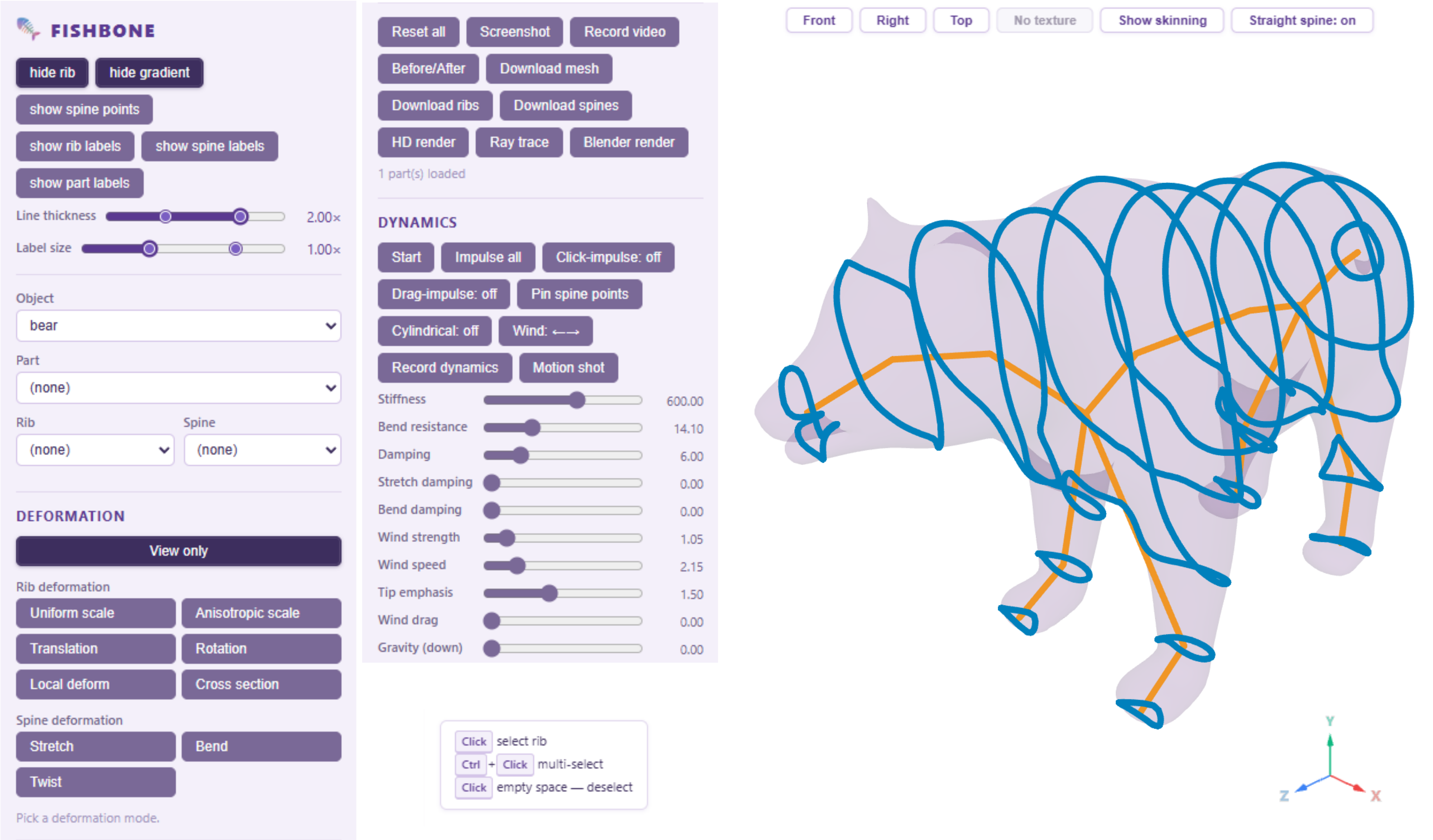}
    \caption{\textbf{Fishbone editor GUI.} The interactive editor displays the input mesh together with its rib polylines and spine. The user picks a rib or spine-branch handle from the side panel and adjusts its parameters with sliders, invoking the deformation primitives of \S\ref{sec:rib_deformation}. Edits are applied at interactive rates via the pre-cached skinning matrix $\mathbf{W}$ (\S\ref{sec:weights_gen}), with no per-edit recomputation.}
    \label{fig:gui}
\end{figure}

\section{Agent Tools}
We list the ten-tool API exposed to the VLM agent of \S\ref{sec:app_agent}, where each call returns text together with a rendered screenshot for closed-loop visual feedback.

\begin{table}[h]
    \centering
    \small
    \begin{tabular*}{\linewidth}{@{\extracolsep{\fill}}ll@{}}
        \toprule
        \textbf{Tool} & \textbf{Purpose} \\
        \midrule
        \texttt{list\_parts}            & Color-coded overview of repo parts \\
        \texttt{set\_part}              & Activate a part \\
        \texttt{list\_ribs}        & Labelled view of all ribs \\
        \texttt{select\_ribs}      & Pick ribs to edit \\
        \texttt{list\_spine\_branches}  & Labelled view of spine branches \\
        \texttt{select\_spine\_branch}  & Pick a branch to edit \\
        \texttt{deform(mode, params)}   & Apply one of nine primitives (\S\ref{sec:rib_deformation}) \\
        \texttt{reset}                  & Revert to rest pose \\
        \texttt{screenshot(view)}       & Re-render from any of six camera angles \\
        \texttt{done(message)}          & Terminate with summary \\
        \bottomrule
    \end{tabular*}
    \caption{\textbf{Agent tool surface.} Tool surface exposed to the LLM agent. All tools return text plus a PNG screenshot, so the agent perceives the mesh visually after every action.}
    \label{tab:agent_tools}
    \vspace{-10pt}
\end{table}

\end{document}